\documentclass{article}
\usepackage{arxiv}
\usepackage[utf8]{inputenc} 
\usepackage[T1]{fontenc}    
\usepackage{hyperref}       
\usepackage{url}            
\usepackage{booktabs}       
\usepackage{amsfonts}       
\usepackage{nicefrac}       
\usepackage{microtype}      
\usepackage{lipsum}		
\usepackage{graphicx}
\usepackage[numbers]{natbib}
\usepackage{doi}
\usepackage[title]{appendix}%

\usepackage{amssymb}
\usepackage{amsmath}

\usepackage{algorithm}
\usepackage{array}
\usepackage[caption=false,font=normalsize,labelfont=sf,textfont=sf]{subfig}
\usepackage{textcomp}
\usepackage{stfloats}
\usepackage{url}
\usepackage{verbatim}
\usepackage{graphicx}
\usepackage{booktabs}
\usepackage{xcolor}

\newcolumntype{H}{>{\setbox0=\hbox\bgroup}c<{\egroup}@{}}

\newcommand{\acronym}{{CMOSFET-Single}}
\newcommand{\acronymm}{{CMOSFET}}
\newcommand{\simcos}{\mathrm{sim}}
\newcommand{\parafango}[1]{\vskip 1mm \textbf{#1.}$\quad$}
\newcommand{\parafanghino}[1]{\vskip 1mm \textit{#1.}$\quad$}

\def\<#1>{[\hbox{$\mkern1mu\thickmuskip=\thinmuskip#1\mkern1mu$}]} 
\definecolor{bluespec}{HTML}{0000ff}
\definecolor{orangespec}{HTML}{FF9900}
\definecolor{magentaspec}{HTML}{ff00ff}
\usepackage{algpseudocode}
\usepackage{multirow}
\usepackage{enumitem}
\newcommand{\mycirc}[1][black]{\Large\textcolor{#1}{\ensuremath\bullet}}

\begin{document}



\title{Continual Learning of Conjugated Visual Representations through Higher-order Motion Flows}


\author{ \href{https://orcid.org/0000-0002-7935-0411}{\includegraphics[scale=0.06]{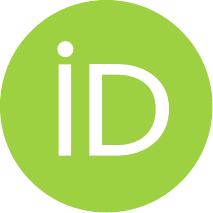}\hspace{1mm}Simone Marullo} \\
	Department of Information Engineering, \\ University of Florence,\\
	Florence, Italy \\
	\texttt{simone.marullo@unifi.it}
\And
\href{https://orcid.org/0000-0002-9133-8669}{\includegraphics[scale=0.06]{orcid.pdf}\hspace{1mm}Matteo Tiezzi} \\
	PAVIS,\\
	Istituto Italiano di Tecnologia (IIT),\\
	Genova, Italy \\
	\texttt{matteo.tiezzi@iit.it}
 \And
\href{https://orcid.org/0000-0001-6337-5430}{\includegraphics[scale=0.06]{orcid.pdf}\hspace{1mm}Marco Gori} \\
	Department
of Information Engineering \\ and Mathematics, University of Siena, \\
	Siena, Italy \\
	\texttt{marco.gori@unisi.it}
 \And
\href{https://orcid.org/0000-0002-0415-0888}{\includegraphics[scale=0.06]{orcid.pdf}\hspace{1mm}Stefano Melacci} \\
	Department
of Information Engineering\\ and Mathematics, University of Siena,\\
	Siena, Italy \\
	\texttt{stefano.melacci@unisi.it}
}

\date{}
\maketitle

\begin{abstract}
Learning with neural networks from a continuous stream of visual information presents several challenges due to the non-i.i.d. nature of the data. However, it also offers novel opportunities to develop representations that are consistent with the information flow. In this paper we investigate the case of unsupervised continual learning of pixel-wise features subject to multiple motion-induced constraints, therefore named \emph{motion-conjugated feature representations}. Differently from existing approaches, motion is not a given signal (either ground-truth or estimated by external modules), but is the outcome of a progressive and autonomous learning process, occurring at various levels of the feature hierarchy. 
Multiple motion flows are estimated with neural networks and characterized by different levels of abstractions, spanning from traditional optical flow to other latent signals originating from higher-level features, hence called higher-order motions. Continuously learning to develop consistent multi-order flows and representations is prone to trivial solutions, which we counteract by introducing a self-supervised contrastive loss, spatially-aware and based on flow-induced similarity. We assess our model on photorealistic synthetic streams and real-world videos, comparing to pre-trained state-of-the art feature extractors (also based on Transformers) and to recent unsupervised learning models, significantly outperforming these alternatives.
\end{abstract}






\keywords{ Learning from Constraints \and Lifelong Learning \and Feature Extraction from a Single Video Stream \and Online Learning.}

\section{Introduction}
\label{sec:intro}


Learning from a continuous stream of visual data is a longstanding challenge for the machine learning community and AI in general. The challenging nature of this task becomes evident when we consider an artificial agent which continuously processes the streamed data, learning in an online manner \cite{ijcaistocazzic}.
The recent outbreak of self-supervised techniques in computer vision has closed the gap with supervised baselines in many vision tasks \cite{moco}, leveraging the intuition of relating different views of the same entity, usually with the goal of building image-level representations, but without explicitly relying on motion.  
Most of the existing approaches are based on contrasting the similarity of positive examples with the dissimilarity of negative examples, using memory banks or large batches \cite{simclr}.
While contrastive learning works surprisingly well for a variety of downstream tasks \cite{surveycont}, typical models require offline training on very large collections of images and some kind of prior knowledge.

{
Our starting point is the observation that unlike machines, humans and animals do not learn by going through vast and randomly shuffled sets of images. Instead, they learn continuously, as they experience the world around them, without storing and labelling every piece of information they come across. In this paper, we argue that machines too can potentially adopt this natural way of learning, acquiring knowledge continuously from their surroundings, without depending on large pre-existing datasets. This would involve, for instance, machines processing visual information as it comes in and occasionally interacting with humans for guidance. Although this goal might seem distant, this work is a small step in that direction. }

The first central point we consider in this paper is the opportunity of substantially relying on motion to design a natural contrastive framework for agents that progressively learn from a visual stream.
Foundational studies in vision and perception have shown \cite{spelke1990principles} that the presence of motion significantly enhances the ability of biological perceptual systems to identify and segment visual patterns into specific entities. Empirical evidence \cite{ostrovsky2009visual} supports the idea that the capability of parsing static visual scene is systematically achieved much later than the one of parsing dynamic scenes. As a matter of fact, the Gestalt Principles of common fate \cite{wertheimer1938laws} hypothesized the role of motion as a fundamental cue for visual perception in the early 20th century. Recently, researchers have applied this concept to the domain of machine learning for computer vision, leveraging motion-based principles to develop the visual skills of artificial agents. Such intuition has been exploited for designing simple pretext tasks in the context of unsupervised learning 
\cite{mahendran2019cross}, aligning the similarity between pairs of feature vectors to the similarity
between the corresponding flow vectors. 
The authors of \cite{pathak2017learning} used segments from low-level motion-based grouping to train convolutional networks, yielding easily transferable representations. The model proposed in \cite{ijcaistocazzic} learns from a video stream, exploiting motion and focus-of-attention provided as external cues. 

In order to develop the aforementioned visual agents, the temporal dimension is of utmost importance, that is the second focal aspect of this paper. Recently, a lot of emphasis has been put on neural models able to learn over time \cite{tiezzi2020focus,tiezzi2022foveated}, in the attempt of overcoming the conventional i.i.d. assumption and offline learning (i.e.,  all training data are  instantly available and
sampled from a static distribution in an independent manner). Most of the attention has been focused on continual supervised learning, with few notable 
unsupervised exceptions \cite{madaan2022representational,rao_razvan,betti2022continual}. Multiple families of approaches have emerged \cite{gidonew}, for which the reader can find a comprehensive description in \cite{tinnesurvey}. Despite the large variety of proposals, learning over time in a continual manner is still a very challenging learning setting, especially when not pushing the emphasis on the memorization of large buffers of past experiences. Regularization techniques are indeed very powerful, and motion seems to offer a natural way of regularizing over time.

Motivated by the aforementioned considerations, we propose to face the learning problem with a  
 neural architecture designed to jointly learn to estimate motion and extract motion-conjugated features (representations) \cite{deeplearningtosee} in a continual manner, which we refer to as 
Continual MOtion-based Self-supervised Feature ExTractor (\acronymm{}). 
Motion can be the outcome of virtually tracking low-level representations (pixel intensities) or higher level features computed by the deep architecture, thus we go beyond the specific notion of optical flow and we introduce higher-order motion flows.
We build on the experience of \cite{ijcaistocazzic} and we propose a more sophisticated self-supervised motion-driven contrastive criterion over time and space, 
learning to estimate multiple motion fields from scratch, without any additional external inputs.  
In the proposed framework, not only optical flow is estimated in a continual learning fashion, as in \cite{marullo2022continual}, but also higher-order flows are continuously estimated, computed according to features extracted at several levels in the feature hierarchy. The idea of exploiting multiple motion cues to guide or support learning is the subject of recent theoretical studies \cite{deeplearningtosee,betti2021visual}.

Our approach processes frames in an online manner with a two-branch neural architecture, continuously-and-jointly learning to extract pixel-wise motion fields and pixel-wise visual features, in a self-supervised manner. Our contrastive criterion revisits the definition of positive and negative pairs, establishing spatio-temporal correspondences/contrasts among pixels belonging to consecutive frames. We make learning affordable by restricting the contrastive loss to a group of stochastically selected pixels, proposing to bias the sampling procedure by the way motion and features are predicted over the frame area.
Motion coherence naturally regularizes feature learning over time, and we gain further stability and reduce catastrophic forgetting by introducing a fast learner / slow learner implementation, where the slow learner model smoothly adapts to the video stream via a momentum-based update scheme.
Our experiments on a recently introduced benchmark and real-world videos confirm the validity of \acronymm{}, and the usefulness of higher-order motion flows. \acronymm{} outperforms the most related models and shows competitive results when compared to convolutional and Transformer-based networks, even if pre-trained on large collections of images.

In this paper, which is an extension of our previous manuscript \cite{ecai2024}, we introduce a layered architecture where both visual features and motion estimation are extracted at multiple levels of abstraction. We further propose a mechanism tailored to tightly couple the development of features and motion estimation with cross-layer relations. We experimentally compare our previous proposal \cite{ecai2024}, hereinafter referred to as \acronym{}, with the \acronymm{} approach presented in this paper and demonstrate that the latter shows significant improvements.
%
Our work is organized as follows. {Section \ref{sec:related} draws connections with related works.} Section~\ref{sec:model} describes the proposed model. Different features of what we propose are described in multiple subsections within Section~\ref{sec:model}. Experiments and the results we obtained are discussed in  Section~\ref{sec:exp}, joinly with the main limitations of what we propose. Finally, conclusions are drawn in Section~\ref{sec:conclusions}, with suggestions for future work.



\section{Related Work}\label{sec:related}

This work illustrates \acronymm{}, a novel method aimed at developing unsupervised visual representations by exploiting motion cues as the main signal for a contrastive learning scheme, designed in a continual learning setting. In the following, we briefly discuss the positioning of the proposed method with respect to such literature domains.

\parafango{Continual Learning} 
Recently, a lot of emphasis has been put on neural models able to learn over time \cite{tiezzi2024continual}, with the aim of eliminating the conventional i.i.d. assumption and offline learning (i.e.,  all training data is available from the start and
sampled from a static distribution in an independent manner). Most of the attention has been focused on continual supervised learning, with few notable 
unsupervised exceptions \cite{madaan2022representational,rao_razvan}. 
Multiple families of approaches have emerged \cite{gidonew}, namely context-specific components, parameter regularization, functional regularization,  replay-based methods. The goal of \acronymm{} is to learn to predict class labels for pixels in a video stream without a traditional offline training procedure. Such goal is not fully defined in advance, since the agent becomes aware of classes as the human supervisor tells him, and it does not go through a sequence of distinct tasks with clear boundaries–being close to CL task-free scenario \cite{aljundi2019task}. Moreover, the spatiotemporal loss functions we propose introduce a motion-induced regularization effect, arguably falling under the umbrella of regularization-based CL. While most of CL literature deals with image classification benchmarks, here we face the challenge of learning from unsupervised long visual experiences unrolling over time.

\parafango{Unsupervised learning by motion cues}
The idea of exploiting motion cues to support learning 
has been exploited for designing pretext tasks in the unsupervised learning community \cite{mahendran2019cross}, aligning the similarity between pairs of feature vectors to the similarity
between corresponding flow vectors 
or using segments \cite{pathak2017learning} from low-level
motion-based grouping to train CNNs.
Tiezzi et al. \cite{ijcaistocazzic} train models on video streams, leveraging a specific motion cue to ensure consistent representations of moving entities along a human-like attention trajectory \cite{tiezzi2020focus}, that is given as an external signal. In contrast, \acronymm{} employs a more sophisticated self-supervised objective that transcends the limitations of a simulated attention trajectory. Remarkably, our approach also learns to estimate the motion field \cite{marullo2022continual} without requiring any additional inputs. Our learning criterion is inspired by the concept of integrating motion prediction with feature extraction, while simultaneously acquiring these skills over time and at multiple levels of abstraction—a novel direction, to the best of our knowledge.
\parafango{Learning in temporally-redundant streams}
The very nature of videos, made of sequences of frames with slowly-varying, temporally-redundant signals, suggests intriguing research directions. Instead of aiming at representing an individual frame, it is possible \cite{deltadist} to estimate the difference between the representations of frames, so to ensure greater temporal consistency in the prediction output and faster processing in a knowledge distillation setting. In our work we do not estimate the feature deltas over time, but rather we assume that the extracted features are consistent when we take into account a motion-induced transformation between the frames.
On a different note, unsupervised development of features from temporally coherent data has been investigated in Slow Feature Analysis (SFA) \cite{wiskott2002slow}, with more
recent applications to high-level tasks, such as action recognition \cite{Sun_2014_CVPR}. The basic idea is to extract features that are slowly varying with respect to the quickly varying input signal, as it is the case for a natural video with moving entities. The concept of slowly varying features is similar to our temporally-consistent features, but in our work we do not impose any strong bias on their structure, they are defined at pixel-level and we explicitly bind their development to motion.

\parafango{Contrastive self-supervised learning (SSL)}
In the context of unsupervised representation learning, unsupervised contrastive methods have recently closed the gap with supervised baselines in many vision tasks \cite{moco}.
Many existing approaches are based on contrasting the similarity of positive examples with the dissimilarity of negative examples \cite{simclr}.
Recent works have replaced negative pairs with  asymmetries in the way features are extracted \cite{byol,simsiam}, with still limited but improving theoretical understanding \cite{simsiam,simsiamhow}. 
In order to increase the stability of the learning dynamics, many works involve an auxiliary model trained by moving average of the main network parameters \cite{moco,byol}, an intuition resumed by \acronymm{}.

The vast majority of the methods can be considered global contrastive learning, since they are explicitly designed to obtain discriminative features for global tasks (e.g., image classification).  For tasks that require local
details (e.g., semantic segmentation), there exist a few recent works on pixel-wise features \cite{flowequivariance,xie2021propagate}, even though they train offline on large scale data or learn in a latent space at smaller resolution, thus not at a truly pixel-wise level. 
%
%
\acronymm{} really embraces this challenge to learn pixel-wise representations from a single long video stream. Moreover, it works in a continual manner, which implies a completely different learning dynamics from traditional SSL methods. Those procedures typically exploit collections of small clips, offline-processed by stochastic optimization \cite{qian2021spatiotemporal}, and mainly for global video representation learning. 
Conversely, in our experiments we exploit the benchmark proposed by the main competitor \cite{ijcaistocazzic}, which has been proposed to mimic  lifelong video stream, showing that the proposed \acronymm{} achieves the best results.

\section{Model}\label{sec:model}
Let us assume the availability of an endless collection of subsequent frames from a visual source, $\{I_t | t>0\}$, originating a video stream $\mathcal{V}$ at the resolution of $W \times H$, with $\mathcal{X}$ the set of valid pixel coordinates for any frame.
Each spatial location $x \in \mathcal{X}$ of $I_t$  can be associated with a feature vector, carrying information about what is present in such a location and its neighborhood. We talk about {\it pixel-wise feature maps} when referring to the collection of such feature vectors. This paper focuses on the challenge of progressively developing a robust feature extractor $\mathcal{F}$ leveraging the information in $\mathcal{V}$, working in a continual online manner, without storing buffers of past data, and without any supervision. The learnt features can then be used to preprocess visual stimuli in the context of different downstream tasks. It is important to remark that frames are continuously streamed at a constant frame rate, without any temporal limits ($t$ could be potentially $\infty$) and they smoothly change over time, thus they are {\it not} independent. Under these conditions, effective learning of useful features over time is not trivial and we will adopt ad-hoc components for the purpose. 
Following the deep and compositional nature of most modern neural architectures for computer vision, we assume that processing the stream $\mathcal{V}$ with model $\mathcal{F}$ results in extracting pixel-wise features at multiple ($N \geq 1$) levels of abstraction, indexed by $\ell \in [1, N]$. 
Another key notion for the purpose of this paper is the one of optical flow, that is widely known in the computer vision literature. Optical flow is the apparent motion associated to the pixels of the visual input, $I_t$, and it can be estimated with a variety of algorithms \cite{brox2004high}. Flow estimation can be efficiently performed by neural networks \cite{flownet}, also when trained in an unsupervised manner \cite{teed2020raft}, 
and a recent study \cite{marullo2022continual} showed that networks for flow extraction can be trained in a continual online fashion, without replay buffers or specific continual learning methodologies. 

In the following we fully describe a model that, for each pixel of the input frame, computes both ($i.$) visual features and ($ii.$) motion estimation, at multiple levels of abstraction. 
Such information is strongly coupled, since the visual feature extractors and the motion estimators are learned in a joint manner. 
In particular, motion is the only driving signal for the development of the visual feature extractor, which conquers equivariance properties naturally induced by the motion fields.
%
%

\subsection{Multi-Order Feature Flows} 
\begin{figure}[!ht]
    \centering
    \includegraphics[width=.6\textwidth]{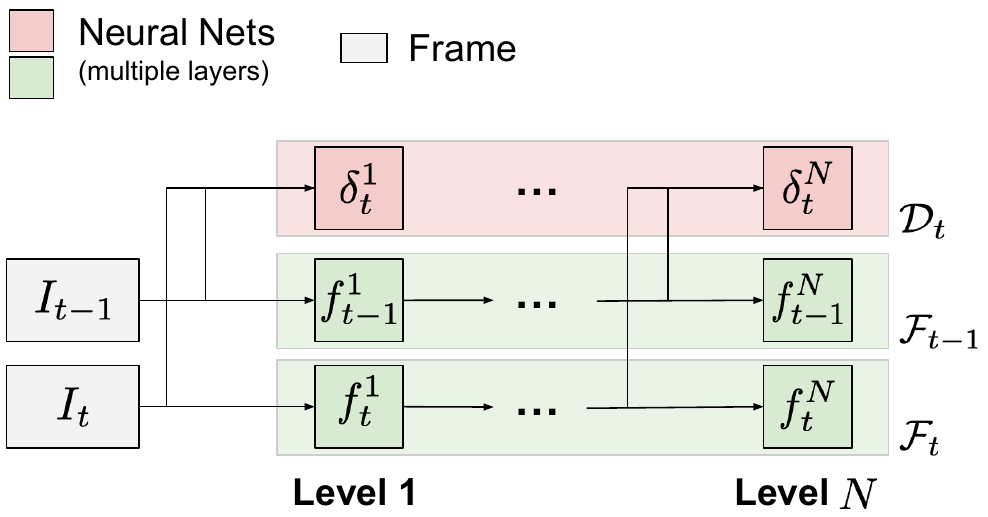}
    \caption{The architecture of \acronymm{}. Given a pair of consecutive frames $I_{t-1}$ and $I_{t}$, pixel-wise features $\bigl(f_{t_1}^{\ell}, f_{t}^{\ell}\bigr)$ and motion flow $\delta_{t}^{\ell}$ are extracted at multiple levels, indexed by $\ell$. }
    \label{fig:toy}
\end{figure}

We indicate with $f^\ell_t$ the feature extractor of the $\ell$-th level, working at time $t$, and consuming the output of $f^{\ell-1}_t$ as input, see Fig.~\ref{fig:toy} (green boxes). 
The notation $f^\ell_t(x)$ indicates the feature vector of length $d^{\ell}$ from level $\ell$ at the 2D spatial coordinates $x$ and $\mathcal{F}_t = \{ f^\ell_t,\ \ell = {1, \ldots L} \}$ with $\ell=0$ being the degenerate case ($f^0_t = I_t$). 
We now introduce the generic notion of \emph{feature flow} 
as a generalization of optical flow when associated to arbitrary visual features $f_t^{\ell}$, including $I_t$ and more abstract features, motivated by the theoretical insights of \cite{deeplearningtosee,betti2021visual}.
The notation $\delta_t^{\ell}$ is used to indicate the function that extracts the $\ell$-th order feature flow, being $\delta_t^{\ell}(x)$ the flow vector at location $x$. 
Here and in the rest of the paper, for the purpose of conciseness, the aforementioned  symbols $f$ and $\delta$ will be overloaded depending on the specific context in which they are exploited, using them to denote functions or to represent their respective outputs, without any ambiguities.

It is convenient to represent motion in terms of displacement vectors between two consecutive frames, so that the feature flow $\delta_t$ establishes a point-wise correspondence between some features at $f_{t-1}$ and $f_{t}$, and it allows the approximate reconstruction $f_{t-1}$ given $f_{t}$ (or vice-versa).\footnote{We removed the layer index on purpose, since this description is generic. 
As it will become clear shortly, multiple feature levels can be associated to the same flow.}
Formally, neglecting the superscript $\ell$:
\begin{equation}
    f_{t-1}(x) \approx W(f_t, \delta_t)(x) := f_t(x+\delta_t(x))  
    \label{eq:warpachetipassa}
\end{equation}
being $W$ the \emph{warping} operator.
The flow $\delta_t$ can be computed by a neural network $\delta$ that processes the representations of a pair of frames, at time $t-1$ and $t$, following \cite{flownet,marullo2022continual}.
We cast this principle into the $N$-levels architecture described so far, so that $\delta^\ell_t$ is the feature flow at level $\ell$, and it exploits the frame representations produced at the layer below, $f_{\cdot}^{\ell-1}$, where we define $f^0_{\cdot}(x) := I_{\cdot}(x)$ for any timestep, i.e., the original pixel representations, see Fig.~\ref{fig:toy} (red boxes).
We use the notation $\mathcal{D}$ to indicate the collection of all the neural networks $\delta^\ell$'s, dedicated to the computation of flows, to distinguish them from the feature extractors in $\mathcal{F}$.
%

The classic optical flow has a clear meaning related to cinematic apparent motion of objects with respect to the camera. 
However, it is largely known that this correspondence loses its strict physical interpretation  when considering non-ideal conditions (e.g., strong occlusions, non-textured objects, changes of lightning, etc.) \cite{horn1981determining}. 
This consideration becomes very important when interpreting the feature flows $\delta^\ell_t$ with $\ell \geq 1$. In deep architectures, higher-level features typically correspond to increasingly abstract representations that emerge from the learning process. As a result, it is challenging to explicitly define expectations for $\delta^\ell_t$. Additionally, higher-level features tend to encode information from a larger neighborhood around a given location in $I_t$ (e.g., receptive fields in convolutional networks), causing local changes in $I_t$ to potentially affect feature vectors associated with distant locations. Consequently, our multi-order flows are not accompanied by predetermined interpretation but rather serve as latent signals that facilitate the transfer of information among motion-connected locations, as we will delve into shortly.

\subsection{Flow-Conjugated Representations} 
Several studies have investigated the idea of acquiring features that align with motion \cite{bregler1997learning,flowequivariance,pathak2017learning}. Recently, \cite{deeplearningtosee} introduced the notion of feature fields which are ``conjugate fields with respect to motion''. This concept is characterized by constraints that specify the connections between a collection of arbitrary pixel-wise characteristics and the motion signal.
Following such a work, we consider the bidirectional {\it conjugation} between features and their respective flows of Eq.~\ref{eq:warpachetipassa} as an essential desirable property for learning purposes. 
On one hand, fulfillment of the approximation in Eq.~\ref{eq:warpachetipassa} can be interpreted as the constraint that enforces flows to be consistent with respect to some ``given features'', as in classic optical flow, where the features are brightness levels or colors. On the other hand, it stimulates ``learnable features'' on consecutive frames to be consistent with some estimated flow, that is the dual of plain optical flow, where features are given (visual input $I_t$).
Formally, we define the following consistency penalty $\mathcal{L}_{c}$ for the constraint of Eq.~\ref{eq:warpachetipassa}, 
\begin{equation}
\mathcal{L}_{c}\left(\delta_{t}, f_{t-1}, f_{t}\right) = \frac{1}{wh}\sum_{x}\rho\bigl(f_{t-1}(x) - \ W(f_t, \delta_t)(x)\bigr),
\label{eq:consistency}
\end{equation}
being $\rho$ a differentiable penalty function, that we selected to be the generalized Charbonnier photometric distance, $\rho(a)=(\|a\|^2 + \epsilon)^{\zeta}$ \cite{charbo}, with $\epsilon=0.001$ and $\zeta=0.5$, as in \cite{marullo2022continual}. 

\parafango{Conjugation Loss} 
The consistency penalty of Eq.~\ref{eq:consistency} represents a generic loss that could be straightforwardly exploited for learning purposes, using the features and motion yielded at each level. However, such implementation would not introduce any explicit cross-layer relation between features and flows (a part from the one arising from the cascaded feature computation). 
We propose to exploit three instances of the penalty in Eq.~\ref{eq:consistency} to set up a loss function that shapes the way features and flows are developed across the model,
\begin{equation}
\begin{aligned}
{L}_{conj}^{\ell} = &
\hskip 1.5mm \mathcal{R}(\delta^\ell_t) \\
&+ \mathcal{L}_{c}({\delta_{t}^{\ell}}, f_{t-1}^{\ell}, f_{t}^{\ell}) & \mycirc[bluespec] \ (i.)\ \ \\
&+\mathcal{L}_{c}({\delta_{t}^{1}}, f_{t-1}^{\ell}, f_{t}^{\ell}) & \mycirc[magentaspec]   \ (ii.)\ \\
&+\mathcal{L}_{c}(\delta_{t}^{\ell}, f_{t-1}^{\ell-1}, f_{t}^{\ell-1}), & \ \ \mycirc[orangespec] \ (iii.)
\label{eq:megaconj}
\end{aligned}
\end{equation}

\begin{figure}[ht]
    \centering
    \includegraphics[width=.7\textwidth]{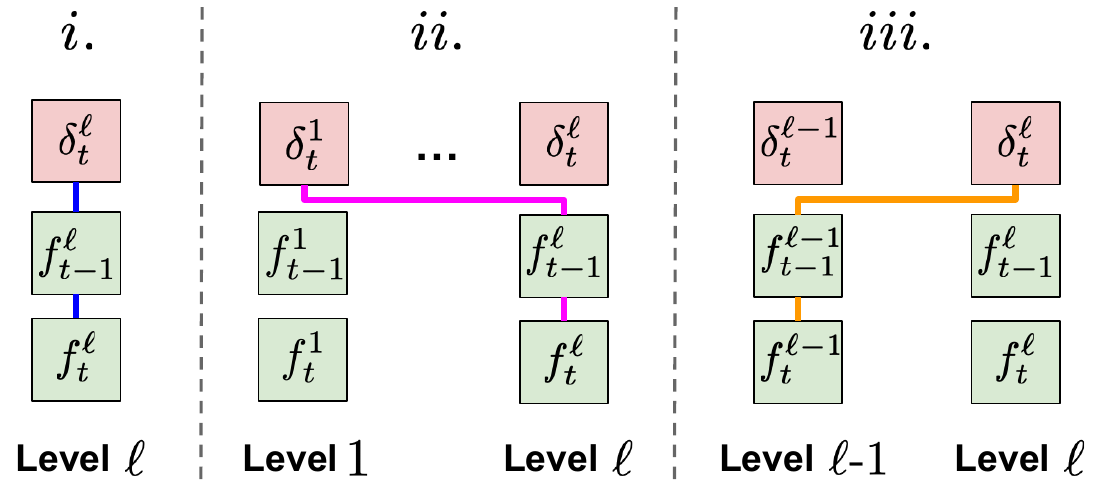}

    \caption{Illustration of the three $\mathcal{L}_c$ terms ($i.$, $ii.$, $iii.$) in Eq. \ref{eq:megaconj}. Each of the three sub-pictures include portions of the architecture of Fig.~\ref{fig:toy}, while connections indicate the dependency introduced by the $\mathcal{L}_c$ term. Case $i.$ is about a single level, while $ii.$ and $iii.$ introduce cross-level dependencies.}
    \label{fig:toy2}
\end{figure}

where $\mathcal{R}(\delta^{\ell}_t)$ is a the usual regularization term of optical flow, proportional to the sum of squares of the spatial gradient of the flow, $\nabla \delta^{\ell}_t$, and required for the well-position of the flow extraction problem \cite{horn1981determining}.\footnote{For the sake of readability, we omit the scaling factors in front of each term in the summation.}
The role of the other three operands in the sum of Eq.~\ref{eq:megaconj} is qualitatively illustrated in Fig.~\ref{fig:toy2}, and described in the following: ($i.$) is the consistency between features and flow computed at layer $\ell$, as expected; 
($ii.$) is the consistency of the features of layer $\ell$ and motion from the first level, encouraging $f^{\ell}$ to be compatible with the displacement yielded by classic optical flow---i.e., learning promotes the preservation of a connection between higher-level features and the lowest-order motion, which is more directly associated with the fine details of the original image; ($iii.$) is the consistency between motion estimated in level $\ell$ with the features of the previous level, to encourage higher-order motion flows to be compatible with less abstract features.  
We notice that the (regularized) term $iii.$ is the usual loss for optical flow estimation applied at the different levels of the feature hierarchy. It is interesting to analyze the meaning of the conjugation loss in the case of $\delta^1_{t}$, the lowest-order flow. Intuitively, this flow is closely related to classic optical flow, since keeping only $iii.$ in Eq.~\ref{eq:megaconj} makes it equivalent to the usual loss for optical flow estimation  (recall that $f^0_{\cdot}$ is $I_{\cdot}$). Moreover, terms $i.$ and $ii.$ degenerate to the same term. Motivated by this considerations, we stop gradients to propagate to $\delta^1_{t}$ through the $i.$ and $ii.$ losses, that allows our model to guarantee that $\delta^1_{t}$ is coherent with the classic optical flow, that also drives the development of $f^{\ell}_{t-1}$ and $f^{\ell}_{t}$ for all the levels of the network. 

\begin{figure*}[ht]
    \centering
    \includegraphics[width=0.9\textwidth]{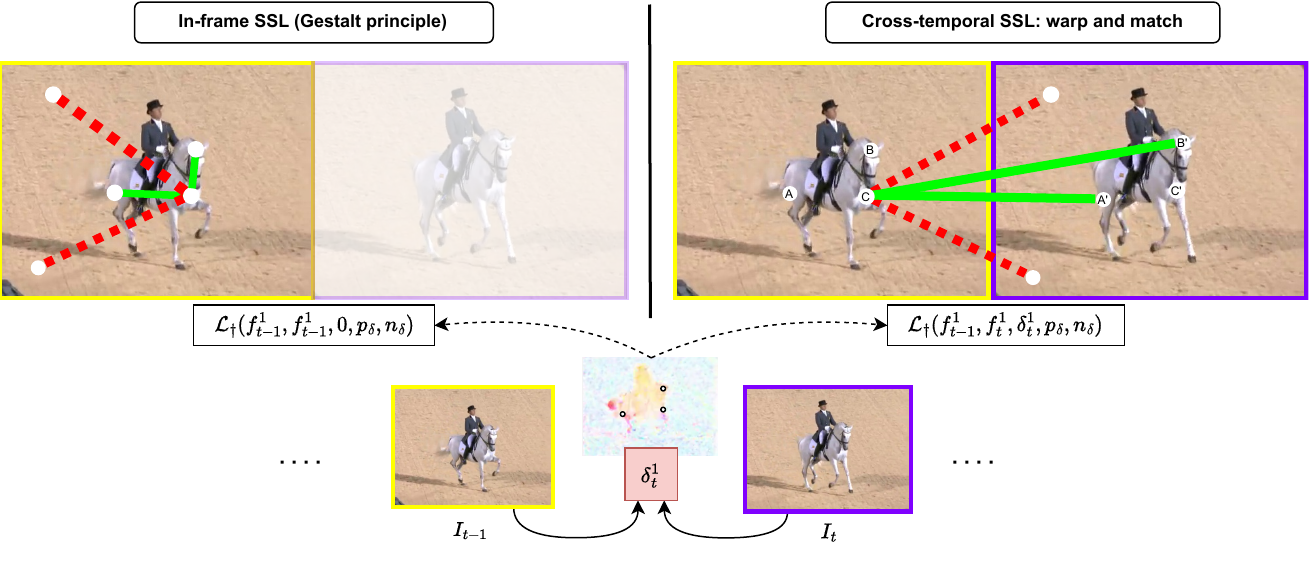}

    \caption{Self-supervised loss, considering a pair of consecutive frames ($I_{t-1}$, $I_{t}$), at the first level ($\ell=1$) of the feature hierarchy (notice that it holds for any $\ell$). The objective $L_{self}^1$ (Eq.~\ref{eq:ssl}) is the sum of two contrastive penalties computed through ${L}_{\dagger}$. The first contribution (left) encourages the development of features by comparing features extracted on the same frame ($I_{t-1}$), while the second term (right) encourages alignment between features extracted in a pair of frames ($I_{t-1}$, $I_{t}$), thanks to flow matching (warping coordinates of items on one side of similarity/dissimilarity relationships, e.g. $A\rightarrow A'$, $B\rightarrow B'$, $C\rightarrow C'$). Green (red) links connect pixels whose features are enforced to be similar (different) according to our motion-based criterion.}
    \label{fig:c}
\end{figure*}

\subsection{Self-Supervised Contrastive Learning} 
Despite these feature-flow constraints, the objective of Eq.~\ref{eq:megaconj} can be easily minimized by trivial solutions (e.g., spatially uniform features with arbitrary flow). Then, we propose to introduce a flow-driven contrastive loss that aligns with the core principles of this paper and encourages the emergence of meaningful features.
Contrastive losses, when minimized, effectively enhance the similarity between features in positive pairs of examples, while concurrently increasing the dissimilarity among features in negative pairs.
Unlike popular contrastive approaches that are aimed at image-level global features \cite{liu2021self,simclr},
 here we focus on the case of pixel-wise representations, thus we reconsider the role of positive and negative examples.  
Our idea consists in following the well-known Gestalt principle which sensibly states that things that move together often carries similar semantic information. 
Furthermore, our objective is to extend this concept to the scenario of learning from a visual stream, where the goal is to establish relationships not only between representations extracted at a single time instant but also across consecutive frames. In order to formally present our contrastive loss, we need to introduce three further ingredients, that are a similarity function $s$, a way to determined positive and negative pairs, a sampling strategy to make learning affordable. We will avoid specifying the layer index $\ell$ unless explicitly needed.
\parafanghino{Similarity Function}
The similarity function $s$ is responsible of comparing features in a generic pair of pixel coordinates $(x_i,x_k)$. Such features are obtained by extractors that we will generically indicate with $g$ and $h$ for the first and second element of the pair, respectively. The reason we introduce two different extractors will be made clear in the following and is related to the learning dynamics. 
In fact, we will consider the case in which $g$ and $h$ operate on the same frame, 
and the case in which they operate on consecutive frames at $t-1$ and $t$. In the former case, no warping of the pixel coordinates will be involved, that is equivalent to consider a degenerate instance of the warping function exploiting $\delta$ that is $0$ for all its inputs. In the latter case, $x_k$ is warped by motion $\delta(x_k)$ before the feature extraction, i.e., we retrieve the location of the warped pixel $x_k$ according to the estimated flow $\delta$, before extracting features.
Formally, $$s(x_i, x_k, \delta, g, h):= \tau^{-1} \simcos(g(x_i),\ h(W(x_k, \delta(x_k))))$$ that is the $\tau$-scaled cosine similarity with $\simcos(a,b):=a\cdot b/(\Vert a\Vert\Vert b\Vert)$ and $\Vert \cdot \Vert$ the Euclidean norm.

\parafanghino{Soft Assignment of Positive and Negative Pairs}
Gestalt assumption \cite{wertheimer1938laws} is exploited to identify positive and negative pairs, considering as positive those pairs of pixels that move coherently, and as negative those pairs that move in a different manner. Of course, this is not expected to be strictly holding in practical scenarios, so we adopt a soft-assignment strategy to mark pairs depending on both distance and flow. 
Given a pixel at $x$, all the spatial coordinates of a given frame are evaluated in order to identify positive and negative examples. As a result, a pair that is composed of $x$ and one of the {\it nearby} pixels with {\it similar motion} is a \emph{positive pair}, while a \emph{negative pair} is composed of $x$ and one of the {\it distant} pixels with {\it different motion}. While this is a sensible learning signal \cite{liu2021self,ijcaistocazzic}, we need to mitigate the effect of the natural violations of the assumption (for example, a non-rigid object will have different motion patterns in different parts of its surface, or there could be a static clone of a moving instance, etc.). 
We indicate with $p_{\delta}(x_i,x_j)$ the confidence score of $(x_i,x_j)$ being a positive pair, while $n_{\delta}(x_i,x_z)$ is the confidence score of $(x_i,x_z)$ being a negative pair, each of them in $[0,1]$. 
Scores $n_{\delta}$ and $p_{\delta}$ are computed as
\begin{equation}
\begin{aligned}
&n_{\delta}(x_i,x_k) = \<\simcos(\delta(x_i), \delta(x_k)) \leq \tau_n> \frac{\| x_i - x_k \|}{\sqrt{H^2+W^2}}, \\  
&p_{\delta}(x_i,x_k) = \<{ \simcos(\delta(x_i), \delta(x_k)) > \tau_p}> \left(1 - \frac{\| x_i - x_k \|}{\sqrt{H^2+W^2}}\right)
 \label{eq:posnegpn}
\end{aligned}
\end{equation}
where $[\cdot]$ is $1$ if the condition in brackets is true, otherwise it is $0$. Thresholds $\tau_{p}, \tau_{n}$ $\in [-1,1]$ are selected to filter out pairs with uncertain similarities, with the condition $\tau_{p} \geq \tau_{n}$ which ensures that $p$ and $n$ are non-zero in a mutually-exclusive manner. The rightmost operands of the products in Eq.~\ref{eq:posnegpn} are the distance between $x_i$ and $x_k$ scaled in $[0,1]$ and $1$ minus such a distance, respectively. Noticeably, motion direction is not significant for static points (motion vector is smaller than a fixed threshold $\tau_{m}$). As such, we set a couple of exceptions to Eq.~\ref{eq:posnegpn}: a moving point and a static point are marked as dissimilar independently on their distance ($n_{\delta}(x_i, x_k) = 1, p_{\delta}(x_i, x_k) = 0$), while a pair of static points is neither similar nor dissimilar ($n_{\delta}(x_i, x_k) = p_{\delta}(x_i, x_k) = 0$). See also the qualitative example of Fig.~\ref{fig:c}, where positive pairs are connected by green lines, while red lines link negative pairs. 

\begin{figure}[t]
  \centering
  \begin{minipage}{1.\textwidth}
  \footnotesize
    \centering
     \fbox{\includegraphics[width=.22\textwidth]{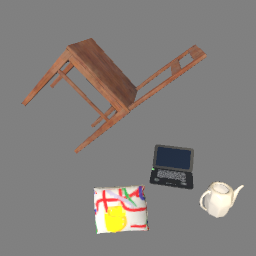}}
     \fbox{\includegraphics[width=.22\textwidth]{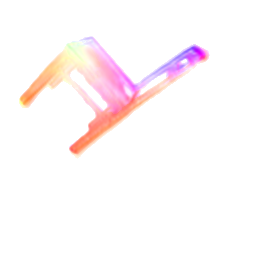}}
     \fbox{\includegraphics[width=.22\textwidth]{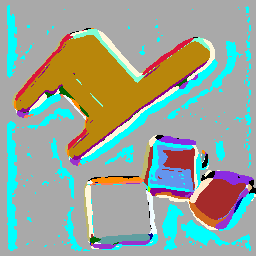}}
     \\
     $\ $ \\
     \vskip -1mm\fbox{\includegraphics[width=.22\textwidth]{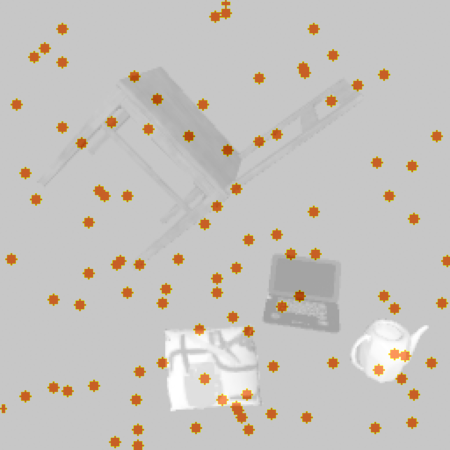}}
     \fbox{\includegraphics[width=.22\textwidth]{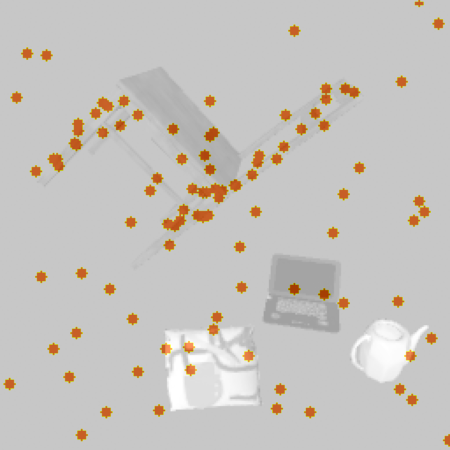}}
     \fbox{\includegraphics[width=.22\textwidth]{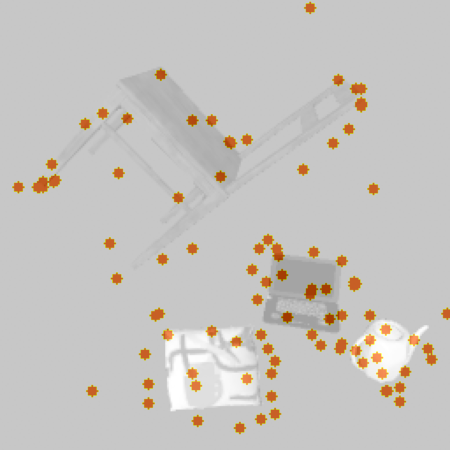}}
     
  \caption{Illustration of different sampling strategies ($\ell=1$). Orange dots (bottom row) are points sampled from a visual stream (top row) in which a chair is moving in a static background that includes three smaller objects (pillow, laptop, teapot). First row: frame, estimated flow (different colors are about different directions), winning feature (different colors are about different winning features). 
  Second row, left-to-right: plain uniform sampling, motion-biased sampling, and the proposed sampling driven by both motion and winning features. In the last case, the sampled coordinates cover both the moving chair and other details of the image in a balanced manner, while the first and second case give more emphasis to the uniform region (first, second) or the moving object (second).} 
  \label{fig:sampling}
  \end{minipage}
\end{figure}

\parafanghino{Sampling}
The number of positive and negative pairs might easily become large, since it is quadratic in the number of pixels, making the evaluation of a constrastive criterion computationally costly.
For this reason, we introduce a motion-driven sampling procedure to make learning affordable. 
This choice is loosely inspired by \cite{ijcaistocazzic}, where the authors proposed to randomly sub-sample the pixel coordinates in function of an externally computed focus of attention trajectory. Differently, we propose a selection criterion based on motion and representations, following the spirit on which our whole proposal is rooted. The outcome of this procedure is a set of $\eta$ coordinates $\mathcal{\tilde{X}}_t \subset \mathcal{X}$, for each $t$, with customizable cardinality $\eta$, on which the proposed contrastive loss will be evaluated.
In detail, with the purpose of injecting priors into the sampling procedure, for any given pair of frames of $\mathcal{V}$ from which we compute features and flows, we model an ad-hoc distribution over the pixel coordinates. Each pixel $x$ at time $t$ has probability $P_{t}(x)$ of being sampled, with $\sum_{x \in \mathcal{X}} P_{t}(x) = 1$. We compute $\{ P_{t}(x), x \in \mathcal{X} \}$ {ensuring that ($i.$) the probability of sampling a point in moving areas is the same as sampling in static areas, and that ($ii.$) the probability of sampling in areas where the $j$-th feature has the strongest activation (absolute value) is the same for all $j$'s. The rationale behind this idea is that ($i.$) we want to ensure that in shots with limited motion the sampling process considers both static and moving areas in a balanced manner (the contrastive loss is built on motion information), and ($ii.$) we want to ensure that different visual patterns are covered in a balanced way by the sampling procedure. In fact, features are naturally associated to (usually) different patterns of the considered input.
Formally, $\mathcal{M}$ is the set of coordinates of moving pixels and $\bar{\mathcal{M}}$ are the coordinates of the static points. }
We compute $\mathcal{S}_j$, $j=1,\ldots,d$, where set $\mathcal{S}_j$ collects the coordinates of the pixels for which the $j$-th component of the learned representation is the largest one, i.e., $\mathcal{S}_j = \{x_z\colon \arg\max | f_{t-1}(x_z) | = j  \}$\footnote{We use $f_{t-1}$ to be consistent with the coordinate frame of $\delta_{t}$, and $|a|$ is the element-wise absolute value of vector $a$.}, where $\arg\max a$ is intended to return the index of vector $a$ associated to the largest component and $|\cdot|$ is the element-wise absolute value. 
It is trivial to see that each pixel coordinate $x$ ends up in belonging to one of the following $2d$ possible sets, $$\{ \mathcal{A}_{j} = \mathcal{S}_j \cap \mathcal{M},\ \mathcal{A}_{2j} = \mathcal{S}_j \cap \bar{ \mathcal{M}},\ j=1,\ldots,d\}.$$ If we let $P_{t}(x)$ return $1 / (2d\cdot\text{card}(\mathcal{A}_{z}))$, being $\mathcal{A}_{z}$ the set to which $x$ belongs and $\text{card}(\mathcal{A}_z)$ its cardinality,\footnote{The overall number of sets ($2d$) can be smaller in the case of one or more empty intersections.},  we get a balanced distribution of the probabilities across the sets,
 $
 \sum_{x \in \mathcal{A}_1} P_{t}(x) = \ldots = \sum_{x \in \mathcal{A}_{2d}} P_{t}(x) = \frac{1}{2d}
$, that fulfills our initial requirements.
Given $P_t(x)$, we sample $\eta > 1$ 
pixel coordinates for each $t$, collecting their coordinates into $\mathcal{\tilde{X}}_{t}$, that is smaller than $\mathcal{X}$. 
In Fig.~\ref{fig:sampling} we report an example of our sampling strategy, comparing uniform sampling with motion-driven sampling and, finally, our motion-and-feature driven sampling. The picture shows how the proposed sampling procedure yields samples that are both related to moving objects (chair) and to portions of the frame with different visual patterns. Differently, a uniform sampling ends up in giving a lot of emphasis to the empty regions of the image.
Fig.~\ref{fig:c} highlights with white circles the elements of $\mathcal{\tilde{X}}_{t}$ (toy example).

\parafango{Contrastive Loss} We are now ready to set up a contrastive loss in the usual log-exponential form \cite{surveycont}, giving more importance to positive pairs with large $p_{\delta}$ and to negative pairs with large $n_{\delta}$,
\begin{equation}
\medmuskip-2mu
\begin{aligned}
&\hskip -1cm {\mathcal{L}}_{\dagger}(g, h, \delta, p_{\delta}, n_{\delta})=\\ &-
\sum_{x_i, x_j \in \mathcal{\tilde{X}}} 
\frac{p_{\delta}(x_i, x_j)}{Z}\log \frac{e^{{s(x_i, x_j, \delta, g, h)}}}{e^{s(x_i, x_j, \delta, g, h)} \ + 
\displaystyle{\sum_{x_z\in \mathcal{\tilde{X}}}}
n_{\delta}(x_i, x_z) e^{s(x_i, x_z, \delta, g, h)}},
\label{eq:fava}
\end{aligned}
\end{equation}
being $Z$ the normalization factor, $Z = \sum_{x, y \in \mathcal{\tilde{X}}_t} p_{\delta}(x, y)$. 
The outer summation is restricted to positive pairs, weighted by their degree of positiveness $p_{\delta}$. The first element of each positive pair, $x_i$, acts as an anchor, so that the similarity of the positive pair is contrasted by all the negative pairs involving $x_i$ (weighted by $n_{\delta}$), as can be appreciated by the denominator of the log-argument.
Notice that the sums of Eq.~\ref{eq:fava} are restricted to the sampled coordinates belonging to $\mathcal{\tilde{X}}_{t}$.
Finally, our self-supervised objective $L_{self}^{\ell}$ is obtained by instantiating Eq.~\ref{eq:fava} twice, relating features from the same frame (at $t-1$)\footnote{This is due to the fact that motion predictors output displacement vectors in the reference frame at $t-1$.} and across two consecutive frames ($t-1$ and $t$), respectively,
\begin{equation}
L_{self}^{\ell} = 
\mathcal{L}_{\dagger}(f^{\ell}_{t-1}, f^{\ell}_{t-1}, 0, p_{\delta}, n_{\delta})+ {\mathcal{L}}_{\dagger}(f^{\ell}_{t-1}, f^{\ell}_{t}, \delta^{\ell}_{t}, p_{\delta}, n_{\delta}).
\label{eq:ssl}
\end{equation}
leading to in-frame contrastive objective and a cross-temporal contrastive objective, respectively. Notice that, in the first term of the summation, the flow is set to $0$, since no warping is performed, as already discussed.
\subsection{Learning Over Time}
\label{sec:continuallearning}
Pairs of consecutive frames are processed in an online fashion, with a single-pass approach for each pair, and without any inputs from auxiliary memory buffers. 

At each time step $t$, a new frame $I_{t}$ is provided by $\mathcal{V}$ and the previous one, $I_{t-1}$, is kept in cache. Learning consists in performing a single update of the model parameters in the whole network ($\mathcal{F}$ and $\mathcal{D}$) given frames $I_{t-1}$ and $I_{t}$, with the goal of minimizing the total loss $L$ that considers both Eq.~\ref{eq:megaconj} and Eq.~\ref{eq:ssl},
\begin{equation}
\begin{aligned}
    {L} = \sum_{\ell}\Bigl( & L_{conj}^{\ell}\bigl(f^{\ell-1}_{t-1}, f^{\ell-1}_t, f^{\ell}_{t-1}, f^{\ell}_t, \delta_t^{\ell}, \delta_t^{1}\bigr) \\ & + L_{self}^{\ell}\bigl(f^{\ell}_{t-1}, f^{\ell}_t, \delta_t^{\ell}\bigr) \Bigr),
    \label{eq:losscum}
\end{aligned}
\end{equation}
where, for completeness, we also made explicit the list of arguments of $L_{conj}^{\ell}$ and $L^{\ell}_{self}$.
Such a loss function $L$ is naturally regularized over time, due to the ${L}_{conj}^{\ell}$ terms in the representation learning criteria, making it a natural instance of regularization-based approaches to continual learning. Moreover, the spatio-temporal bridge introduced on the sampled points by our contrastive loss (second term of the summation in Eq.~\ref{eq:ssl}), introduces further regularity over time. Although it is possible to incorporate replays or other continuous learning techniques \cite{survey} alongside our proposed method (that goes beyond the scope of this paper), this study focuses exclusively on the more demanding yet more authentic scenario of plain continual online learning \cite{ijcaistocazzic}. It is natural to wonder whether such regularity over time is enough to guarantee a good compromise between plasticity and stability of the networks, hence mitigating catastrophic forgetting, in both the motion estimators $\mathcal{D}$ and the feature extractors $\mathcal{F}$.

A recent experience on motion estimation in continual learning settings \cite{marullo2022continual} has shown that the (regularized) criterion of Eq.~\ref{eq:consistency} is enough to learn to estimate pixel displacements $\delta$ over time by online gradient descent. Moreover, forgetting issues are not particularly evident due to the locality of the motion prediction process, resulting in negligible interference even in case of long streams with significant variability. 
%
However, our initial experimental findings indicated that learning over time poses significant challenges for the feature extractors $\mathcal{F}$, as we observed notable forgetting when approaching the problem using the naive approach. Then, we borrow an intuition from models whose weights are updated by a momentum-based moving average 
\cite{emafirstteacher,moco,emaself,emacontinual}. 
Despite being conceptualized in different configurations, these approaches share the idea of using an Exponential Moving Average (EMA) scheme to update the parameters of a (teacher) network, while another network (student), which is continuously updated by gradient descent, is enforced to be consistent with it, preventing rapid changes in the parameter space. The EMA-updated network acts as a slowly progressing encoder, with parameters that evolve smoothly \cite{moco}, and it can be employed both to stabilize learning of the student network and to better preserve information from the past.

\begin{figure}[ht]
    \centering
    \includegraphics[width=0.5\columnwidth]{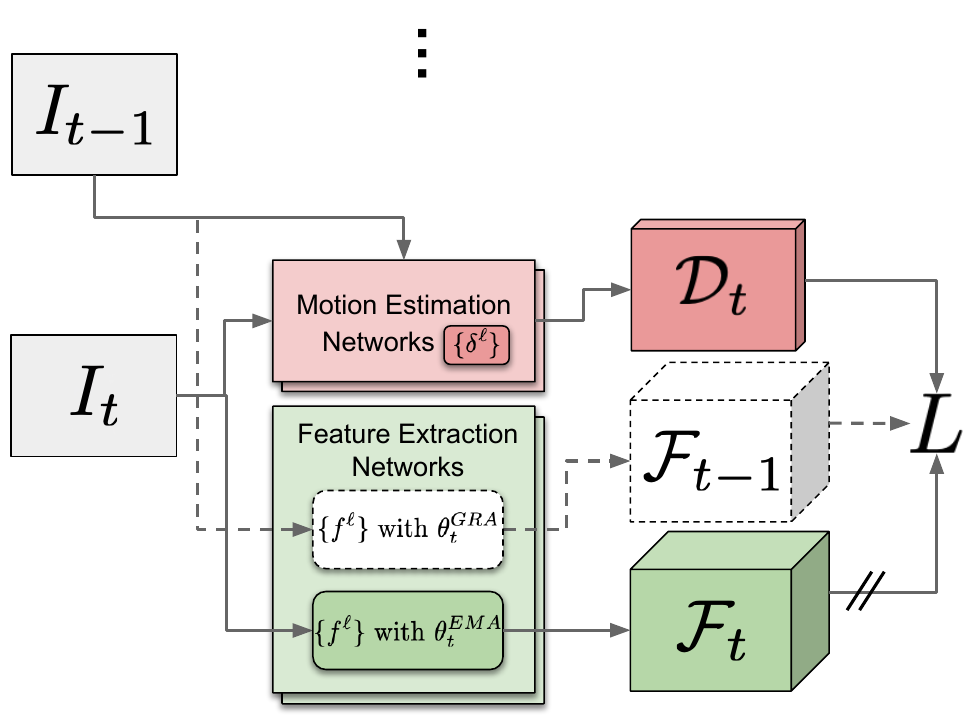}
    \caption{Learning over time with \acronymm{}. Exponential Moving Average (EMA) networks and  gradient-updated networks (GRA) extract features from $I_{t}$ and $I_{t-1}$, respectively. White-dotted parts of the model are in addition to the ones of Fig.~\ref{fig:toy}, and the diagonal bars $\bigl(//\bigr)$ indicate that no gradients are propagated on that path. The loss function ${L}$ of Eq.~\ref{eq:losscum} drives the learning process of the motion predictors and of the feature extractors, enforcing GRA to be coherent with EMA and instantiating the motion-induced contrastive criterion.}
    \label{fig:learning}
\end{figure}

%
Since our model is designed to extract and relate representations from a pair of frames $(I_{t-1}$, $I_{t})$, we propose to extract features from  frame $I_{t-1}$ using a feature extractor whose weights at time $t$, referred to as $\theta_{t}^{GRA}$ (GRAdient-updated), are updated by an online gradient step, while the current frame is processed by an EMA-updated network (slow learner) with the same architecture and weights $\theta_{t}^{EMA}$,
\begin{eqnarray*}
f^{\ell}_{t-1} &=& f^{\ell}(f^{\ell-1}_{t-1}, \theta_{t}^{GRA, \ell})\\
f^{\ell}_{t} &=& f^{\ell}(f^{\ell-1}_{t}, \theta_{t}^{EMA, \ell}),
\end{eqnarray*}
where, for the sake of simplicity, we overloaded the notation used so far making explicit the weight values-related argument.
Fig.~\ref{fig:learning} summarizes the structure of whole model, emphasizing the two types of networks GRA and EMA.
Such two networks are naturally bridged by the ${L}_{self}$ (Eq.~\ref{eq:ssl}) component of the loss function $L$, that enforces coherence between them in the motion-driven manner proposed in this paper. While the GRA network is updated by the gradient of ${L}$ with respect to its weights and using learning rate $\alpha_{f} > 0$, the other network is updated by EMA with coefficient $\xi \in [0, 1)$,
\[
\begin{aligned}
 \theta_{t+1}^{GRA} &= \theta_{t}^{GRA} - \alpha_{f} \nabla_{\theta} {L}(\mathcal{F}_{t-1}, \hat{\mathcal{F}}_{t}, \mathcal{D}_{t})\\
 \theta_{t+1}^{EMA} &= \xi \theta_{t}^{EMA} + (1-\xi) \theta_{t+1}^{GRA},
\end{aligned}
\]
where $\nabla_{\theta}{L}$ is the gradient with respect to the weights in $\mathcal{F}$ and $\hat{\mathcal{F}}_{t}$ indicates that $\mathcal{F}_{t}$ is treated here as a constant value (it comes from the EMA instance).
The learning procedure is summarized in Algorithm~\ref{algo}.

\begin{algorithm}[ht]
    \begin{algorithmic}
    \Require Stream $\mathcal{V}$ of length $T$, potentially infinite; neural nets $\mathcal{D}$ and $\mathcal{F}$ with initial parameters set to $\gamma_{1}$ and $\theta_{1}$. Learning rates $\alpha_{m}$, $\alpha_{f} > 0$ and $\xi \in [0, 1)$.
    \State \vspace{-3mm}$\ $
    \State $t \gets 1,\ \ $ $\theta_{1}^{GRA} \gets \theta_{1},\ \ $ $\theta_{1}^{EMA} \gets \theta_{1}$
    \State $f_0^0 = I_1,\ \ f_1^0 = I_1$
    \State \vspace{-3mm}$\ $
    \While{$t \leq T$}
        \State {$I_{t} \gets$ next\_frame($\mathcal{V}$)}
        \State \vspace{-3mm}$\ $
        \For{$\ell=1, \dots, \ell=N$} 
        \State $\delta^{\ell}_{t} = \delta^{\ell}(f^{\ell-1}_{t}, f^{\ell-1}_{t-1}, \gamma_{t}^{\ell})$
        \State $f^{\ell}_{t-1} = f^{\ell}(f^{\ell-1}_{t-1}, \theta_{t}^{GRA, \ell})$
        \State $f^{\ell}_{t} = f^{\ell}(f^{\ell-1}_{t}, \theta_{t}^{EMA, \ell})$
        \EndFor
        \State \vspace{-3mm}$\ $
        \State $\mathcal{D}_{t} \gets \{\delta_{t}^{\ell} | \ell = 1, 2, \ldots, N\}$
        \State $\mathcal{F}_{t-1} \gets \{f_{t-1}^{\ell} | \ell = 1, 2, \ldots, N\}$  
        \State $\mathcal{F}_{t} \gets \{f_{t}^{\ell} | \ell = 1, 2, \ldots, N\}$ 
        \State \vspace{-3mm}$\ $
        \State $\gamma_{t+1} \gets \gamma_{t} - \alpha_{m} \nabla_{\gamma} {L}(\mathcal{F}_{t-1}, \mathcal{F}_{t}, \mathcal{D}_{t})$     
        \State $\theta_{t+1}^{GRA} \gets \theta_{t}^{GRA} - \alpha_{f} \nabla_{\theta} {L}(\mathcal{F}_{t-1}, \mathcal{F}_{t}, \mathcal{D}_{t})$
        \State $\theta_{t+1}^{EMA} \gets \xi \theta_{t}^{EMA} + (1-\xi) \theta_{t+1}^{GRA}$
        \State \vspace{-3mm}$\ $
        \State $t \gets t + 1$
    \EndWhile
    \end{algorithmic}
    \caption{The learning procedure. For each frame from stream $\mathcal{V}$, features are extracted, motion is estimated, and the proposed self-supervised loss $L$ drives the learning processes. The GRA and EMA networks are initialized to the same weight values.}
    \label{algo}
\end{algorithm}

\section{Experiments}
\label{sec:exp}

\setlength{\fboxsep}{0pt}%
\setlength{\fboxrule}{1pt}%

The proposed framework, \acronymm{}, has been developed\footnote{ \url{https://github.com/sailab-code/unsupervised-learning-feature-flow}}.  using the PyTorch library, running experiments on two Linux machines equipped with an NVIDIA Tesla V100 GPU, 32 GB of memory (see Appendix for further details). The primary objective of this work is to explore the potential of \acronymm{} in generating pixel-wise representations, learning in a continual, online, self-supervised manner, concurrently developing multiple flows estimators.  Our experimental evaluation follows a recently proposed benchmark that was proposed in \cite{ijcaistocazzic} and also inherited by further works \cite{ecai2024}, specifically designed for evaluating the quality of pixel-wise features generated in a continual learning paradigm, that we summarize in the following, after having introduced the datasets we selected. 

\parafango{Datasets} Our proposed method, \acronymm{} is evaluated on two distinct collections of video streams. The first collection, based on the work of \cite{ijcaistocazzic}, comprises three $256\times256$ streams generated from the 3D Virtual Environment SAILenv \cite{DBLP:conf/icpr/MeloniPTGM20,meloni2021evaluating}. 
Video streams are about some target objects moving around and performing complex transformations consisting of rotations, scaling, and pose variations with respect to a fixed camera. The first stream, \textsc{EmptySpace}, depicts a room with uniform background and four moving objects, including a chair, laptop, pillow, and ewer. It includes both grayscale (\textsc{-bw} suffix) and RGB (\textsc{-rgb}) data. The second stream, \textsc{Solid}, comprises three white 3D solids, including a cube, a cylinder, and a sphere, moving in a grayscale environment. This stream requires the developed features to discern context/shapes rather than brightness or color. The third stream, \textsc{LivingRoom} (\textsc{-bw}, \textsc{-rgb}), depicts the same four objects as in \textsc{EmptySpace}, moving around in a more complex scene with a heterogeneous background composed of non-target objects such as a couch, tables, staircase, door, and floor, as well as static copies of the \textsc{EmptySpace} objects. We inherit from \cite{ecai2024} the second collection of video streams, which include two real-world videos, \textsc{Rat} ($256\times128$) and \textsc{Horse} ($256\times192$), proposed by \cite{longvideos}, that recorded the behaviour of rat and a horse (actually horse+jockey), respectively, representing the target objects of the streams. These videos\footnote{ \url{https://www.kaggle.com/datasets/gvclsu/long-videos}} are used to demonstrate \acronymm{}'s ability to generalize to real-world scenarios, with non-fixed camera.

\parafango{Experimental Setup} 
In order to evaluate the performance of \acronymm{}, we conducted a comparative study with the main competitor, the model proposed in \cite{ijcaistocazzic}, on a pixel-wise classification task. Our experiments strictly adhere to the benchmark framework of \cite{ijcaistocazzic}, which consists of unsupervised learning for $30$ laps\footnote{The agent observes the scene from a fixed position, while individual objects of interest move one at a time along smooth trajectories. These objects rotate and change their distance from the camera, either moving closer or farther away. A complete route made by each object, returning to its starting point, is referred to as a \textit{lap}. Please refer to \cite{ijcaistocazzic} for further details on the experimental framework.} per object, with the last $5$ laps involving the storage of $3$ supervised representations (pixel-wise templates) per object, at intervals of $100$ frames, in a small memory referred to as $\mathcal{C}$. 
These externally provided cues are only utilized for an evaluation (classification) procedure, which has no bearing on the representation learning process. They simulate the outcome of a tiny set of interactions with a supervisor, as is typical of realistic learning settings.
The evaluation stage consists in processing the stream for one more lap, not only extracting features but also providing them to an open-set classifier $c(\{f^{\ell}_{t}(x) | \ell = 1, \ldots, N\}, \mathcal{C})$ that predicts the class-membership scores of pixel coordinates $x$ given the representations produced by feature extractors in $\mathcal{F}$ and the templates stored in the memory $\mathcal{C}$. Scores are about a certain number of classes that depends on the considered stream. 
The quality of the predictions of $c$ is measured by the F1-score. The classifier $c$ is implemented as a distance-based model (cosine similarity in our case)\footnote{Features $f^{\ell}_t$ are independently normalized for each $\ell$, in order to account for large differences in magnitude between the extractors.}. In such a way, the model is capable not only to predict the target classes, but also to abstain\footnote{{Open-set classifiers \cite{scheirer2012toward}, in addition to being capable to distinguish between examples belonging to different training classes, can also detect whether data do not belong to any known class, a realistic condition for autonomous visual agents. \acronymm{} is also class-incremental \cite{9040673}, due to the progressive inclusion of new classes after human intervention.}} from prediction when the minimum distance from all the templates populating $\mathcal{C}$ is greater than a certain threshold $\xi$.
In the case of \acronymm{}, we apply a simple learning schedule to warm-up the motion estimators, avoiding premature training of the neural feature extractors. During the first $T_{sched}$ 
laps we only learn the flow estimator at level $1$. Then, if $N > 1$, we start learning the feature extractor on the same level, and we wait further $T_{sched}$ laps before applying the same schedule on the following level. The procedure is repeated, progressively activating learning on  the whole network (see ~\ref{sec:params} for details). 

Following the evaluation protocol of \cite{ijcaistocazzic,ecai2024}, the performance of our approach is evaluated by computing the F1 score (averaged over the available object categories and the background class) over all pixels of the frames, in a last additional lap per object. 
We adopted the exact same methodology to evaluate the performance of our approach on the considered real-world videos. In this case, ground truth for evaluation comes from manual labelling provided by \cite{longvideos} on selected frames. 
Still following the procedure defined in the evaluation protocol of \cite{ijcaistocazzic,ecai2024}, the optimal values for the hyperparameters are chosen by maximizing the F1 along the trajectory (1 pixel per frame) of a given human-like attention model \cite{zanca2020gravitational}. 
The grids of parameter values that we considered and the optimal values are reported in ~\ref{sec:params}, together with further details.

\parafango{Neural Architectures} 
We now present the neural architectures used to implement the feature extractors $\mathcal{F}$ and flow estimators $\mathcal{D}$. Each $f^{\ell}$ and each $\delta^{\ell}$ consists of a \textsc{ResUnet} architecture \cite{ijcaistocazzic}, which is a variant of the U-Net network \cite{ronneberger2015u}. We reduced the number of convolutional filters in each layer by a factor of 4, while the number of per-pixel outputs is 32 for $f^{\ell}$ and 2 for $\delta^{\ell}$ (horizontal and vertical flow components). We considered two implementations of this model: \acronym{}, with a single level of feature and flow extractors (i.e., $N=1$), and \acronymm{}, with two levels, developing lower and higher-order motions (i.e., $N=2$). We compared with the competitors presented in \cite{ijcaistocazzic}: ResUnet and FCN-ND (which has 6 convolutional layers and no downsampling). For more details refer to ~\ref{sec:models}.
%
%
%
As a rough baseline for comparison, we also evaluated the performance of several  models that were pre-trained and specialized in semantic segmentation tasks and self-supervised learning, of course without attempting to outperform their results. Specifically, we evaluated the ResNet101-based \textsc{DeepLabV3} \cite{chen2017deeplab} and the Dense Prediction Transformer - \textsc{DPT} \cite{ranftl2021vision}, both of which have backbones trained on ImageNet-1M (\textsc{-B} suffix) and classification heads (\textsc{-C} suffix) trained on COCO \cite{lin2014microsoft} and ADE20k \cite{zhou2019semantic}. We also considered the pre-trained features from the backbones of recent self-supervised approaches, including \textsc{MoCo v1, v2, v3} \cite{moco,chen2020improved,chen2021mocov3} and \textsc{PixPro} \cite{xie2021propagate}. We upsampled the features picked at one of the last three residual stages (testing all of them) and reported the best-performing configuration. Additionally, we used the dummy baseline \textsc{Raw Image}, which uses the original pixel representations (i.e., pixel color/brightness) as features.

\parafango{Quantitative Results} Table~\ref{tab:main_frame2} reports the evaluation results of our proposed method, considering the F1 scores over the entire frame.
\begin{table*}[t]
\caption{F1 scores over 10 runs  (mean $\pm$ std), in seven different video streams (\textsc{EmptySpace}, \textsc{Solid}, \textsc{LivingRoom}, \textsc{Rat} and \textsc{Horse} with some \textsc{-BW} variants). The bottom part of the table is about the main competitors of the proposed model, including the raw-image (degenerate) baseline. The top part of the table collects {reference} results obtained with large-offline-pre-trained models, publicly available (with no attempts to overcome them).}
\label{tab:main_frame2}
\centering
\small
\scalebox{0.95}
{
\begin{tabular}{cl|Hc|lllllll}
\toprule
& & \multirow{2}{*}{Reference} & \multirow{2}{*}{\# Params} & \multicolumn{2}{l}{\textsc{EmptySpace}} &  \textsc{Solid}  & \multicolumn{2}{l}{\textsc{LivingRoom}} & RAT & HORSE\\ 
 & & & & {\tiny BW} & {\tiny RGB} & {\tiny BW} & {\tiny BW} & {\tiny RGB} & {\tiny RGB} & {\tiny RGB} \\
\midrule
\noalign{\smallskip}
\multirow{9}{*}{\rotatebox{90}{\scriptsize\textsc{$\ \ \ $Offline  Pre-Train.}}}& \textsc{DPT-C} \cite{ranftl2021vision}&  & 121M  &$0.66$&$0.67$&$0.64$&$0.35$&$0.39$ & $0.59$ & $0.83$\\
& \textsc{DPT-B} \cite{ranftl2021vision}& & 120M &${ 0.71 }$&$0.69$&${ 0.68 } $&${ 0.39 } $&$0.39$ & $0.58$ & $0.87$\\
& \textsc{DeepLab-C} \cite{chen2017deeplab}& &58.6M&$0.49$&$0.61$&$0.57$&$0.31$&$0.34$& $0.56$ & $0.86$\\
& \textsc{DeepLab-B} \cite{chen2017deeplab}& &42.5M&$0.70$&$0.65$&$0.66$&$0.34$&${0.44} $& 0.57 & $0.81$\\
& \textsc{MoCo v1} \cite{moco}&& 8.5M &$0.73$&$0.73$&$0.74$&$0.33$&$0.35$& $0.70$ & $0.66$\\
& \textsc{MoCo v2} \cite{chen2020improved}& & 8.5M &$0.75$&$0.76$&$0.74$&$0.41$&$0.43$& $0.59$ & $0.79$\\
& \textsc{MoCo v3} \cite{chen2021mocov3}& & 8.5M  &$0.76$&$0.76$&$0.76$&$0.41$&$0.44$& $0.58$ & $0.64$\\
& \textsc{PixPro} \cite{xie2021propagate}&& 8.5M &$0.31$&$0.46$&$0.41$&$0.30$&$0.22$& $0.59$ & $0.70$\\
\noalign{\smallskip}
\midrule
\noalign{\smallskip}
\multirow{5}{*}{\rotatebox{90}{{\textsc{\color{black}\scriptsize Continual}}}}
& \textsc{Raw Image}& &-&$0.50$&$0.45$&$0.18$&$0.10$&$0.23$& 0.67 & $0.66$\\
&\textsc{HourGlass} \cite{ijcaistocazzic}&& 17.8M&$0.55$\scalebox{.7}{$\pm 0.03$}&$0.71$\scalebox{.7}{$\pm 0.03$}&$0.50$\scalebox{.7}{$\pm 0.01$}&$0.31$\scalebox{.7}{$\pm 0.04$}&$0.25$\scalebox{.7}{$\pm 0.07$}& {$0.59$\scalebox{.7}{$\pm 0.04$}} & {$0.71$\scalebox{.7}{$\pm 0.07$}}\\
& \textsc{FCN-ND}\cite{ijcaistocazzic}&&0.1M &${0.60}$\scalebox{.7}{$\pm 0.05$}&$0.51$\scalebox{.7}{$\pm 0.07$}&$0.48$\scalebox{.7}{$\pm 0.03$}&{$0.24$\scalebox{.7}{$\pm 0.01$}}&{$0.28$\scalebox{.7}{$\pm 0.03$}}& {$0.42$\scalebox{.7}{$\pm 0.05$}} & {$0.50$\scalebox{.7}{$\pm 0.08$}} \\
& \acronym{} & & 1.1M &$0.60$\scalebox{.7}{$\pm 0.06$}&${0.78}$\scalebox{.7}{$\pm 0.06$}&${0.62}$\scalebox{.7}{$\pm 0.02$}&{${0.33}$\scalebox{.7}{$\pm 0.03$}}&{${0.34}$\scalebox{.7}{$\pm 0.05$}}& ${0.61}$\scalebox{.7}{$\pm 0.07$} & ${0.75}$\scalebox{.7}{$\pm 0.06$} \\
& \acronymm{} & & 2.3M &$\textbf{0.64}$\scalebox{.7}{$\pm 0.06$}&$\textbf{0.82}$\scalebox{.7}{$\pm 0.02$}&$\textbf{0.64}$\scalebox{.7}{$\pm 0.05$}&{$\textbf{0.38}$\scalebox{.7}{$\pm 0.04$}}&{$\textbf{0.36}$\scalebox{.7}{$\pm 0.05$}}& {$\textbf{0.74}$\scalebox{.7}{$\pm 0.09$}} & $\textbf{0.84}$\scalebox{.7}{$\pm 0.01$} \\
\bottomrule 
\end{tabular}}
\end{table*} 
Specifically, we focus on the related \emph{continual} self-supervised competitors (bottom part of the table),
observing that \acronymm{} outperforms them in all  video streams. The smaller variant \acronym{} is already slightly superior to the competitors, except for \textsc{EmptySpace-bw}, where it is on par with the best competitor. Notably, the largest gap between \acronym{} and \acronymm{} is observed in real-world videos (\textsc{Rat}, \textsc{Horse}) that confirms the importance of higher-order motion. We remark that in our proposed approach, the flow estimation is learned online from scratch, in contrast to the competitors, that rely on a pre-computed motion signal that is flawlessly produced by the rendering engine. Moreover, \acronymm{} methods do not rely on explicit segmentation procedures for learning, yet they effectively exploits the relationships between flows and learned features. Comparing \textsc{EmptySpace-bw} and \textsc{EmptySpace-rgb}, we notice a significant gap in F1, suggesting that even though the RAW scores of grayscale and color representations are not so different, both \acronymm{} and the competitors seem to efficiently exploit the intrinsically richer information encoded in RGB.
Additionally, we observe that \acronymm{} encodes pixels in a reduced number of output features compared to competitors ($32\cdot 2$ vs. up-to-128), and has significantly fewer learnable parameters than the best competitor (on average), with 2.3M parameters compared to 17.8M (third column of Tab.~\ref{tab:main_frame2}). This confirms the capability of our method to develop informed but more compact representations. Surprisingly, our approach achieves interesting performance even when compared to offline-pre-trained large architectures with several millions of parameters (upper part of the table). It is important to note that our goal is not to overcome large-scale massively offline-trained models, since we learn from scratch in an online manner, specializing in the target environment. Nonetheless, \acronymm{} beats these competitors in \textsc{EmptySpace-rgb} and \textsc{Rat-rgb}. We highlight the performance on the other real-world video stream, \textsc{Horse-rgb}, where \acronymm{} overcomes most of the offline-trained models (all of the self-supervised ones) and it is only slightly behind the best one. These results affirm the adaptability of our approach to the properties of the environment at hand, with a fraction of the learnable parameters and processed data.

\begin{figure*}
  \centering
  \begin{minipage}{1.0\textwidth}
  \footnotesize
    \centering
     \hskip 0.5mm \rotatebox{90}{\hskip 3.5mm \textsc{EmptySpace}} 
     \fbox{\includegraphics[width=.19\textwidth]{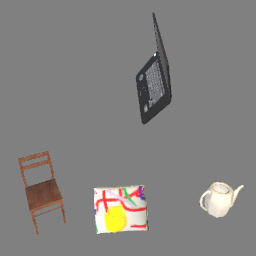}}
     \fbox{\includegraphics[width=.19\textwidth]{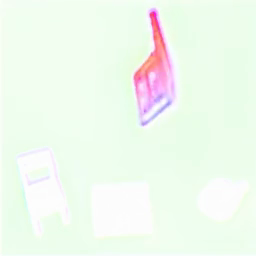}}
     \fbox{\includegraphics[width=.19\textwidth]{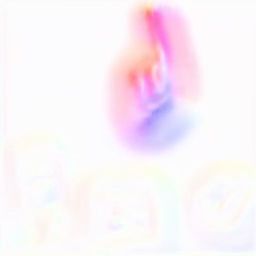}}
     \fbox{\includegraphics[width=.19\textwidth]{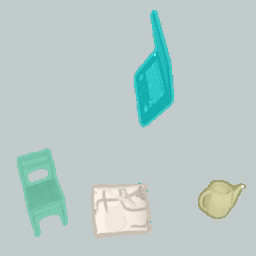}}
     \raisebox{.40\height}{\includegraphics[width=.12\textwidth,trim=2mm 12mm 1mm 1mm,clip]{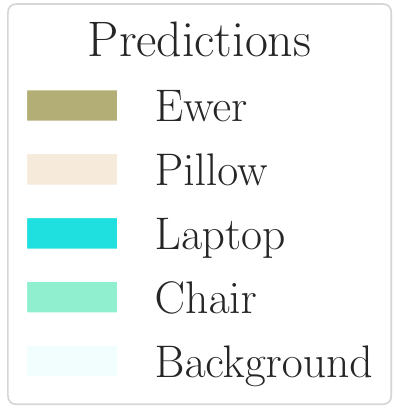}}  
     \\
     \vskip 1mm\rotatebox{90}{\hskip 2.5mm \textsc{Rat}}
     \fbox{\includegraphics[width=.19\textwidth]{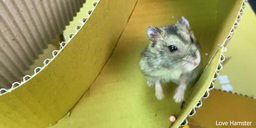}}
     \fbox{\includegraphics[width=.19\textwidth]{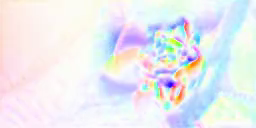}}
     \fbox{\includegraphics[width=.19\textwidth]{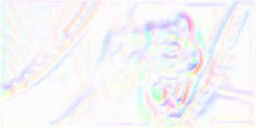}}
     \fbox{\includegraphics[width=.19\textwidth]{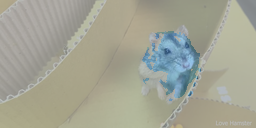}}
     \hphantom{\raisebox{.27\height}{\includegraphics[width=.1175\textwidth,trim=2mm 12mm 1mm 1mm,clip]{fig/seg_legend.pdf}}}
     \\
     \vskip 1mm
     \hskip 0.05mm \rotatebox{90}{\hskip 4mm \textsc{Horse}}
     \fbox{\includegraphics[width=.19\textwidth]{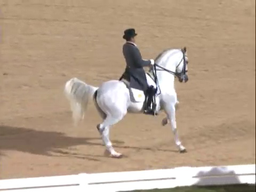}}
     \fbox{\includegraphics[width=.19\textwidth]{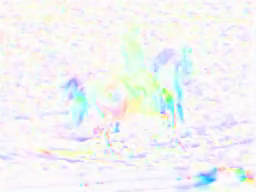}}
     \fbox{\includegraphics[width=.19\textwidth]{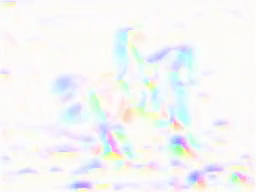}}
     \fbox{\includegraphics[width=.19\textwidth]{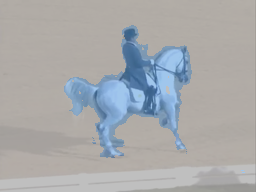}}
     \hphantom{\raisebox{.27\height}{\includegraphics[width=.1175\textwidth,trim=2mm 12mm 1mm 1mm,clip]{fig/seg_legend.pdf}}}
     
  \end{minipage}

  \begin{minipage}{0.8\textwidth}
  \caption{Three frames sampled from \textsc{EmptySpace}, \textsc{Rat} and \textsc{Horse} streams (first column). 
  \acronymm{} flows (first-order and higher-order in second and third columns, respectively) are plotted according to the optical flow conventional color mapping \cite{colorof} (see the color wheel on the right). Predictions in the fourth column show a pretty satisfying adherence to the borders of the objects (target objects are highlighted with colors; the background class is present in {\it all} the streams, reported with a light gray overlay).} 
  \label{fig:main_frame}
  \end{minipage}
  $\ $
  \begin{minipage}{0.125\textwidth}
      {\includegraphics[width=1\textwidth]{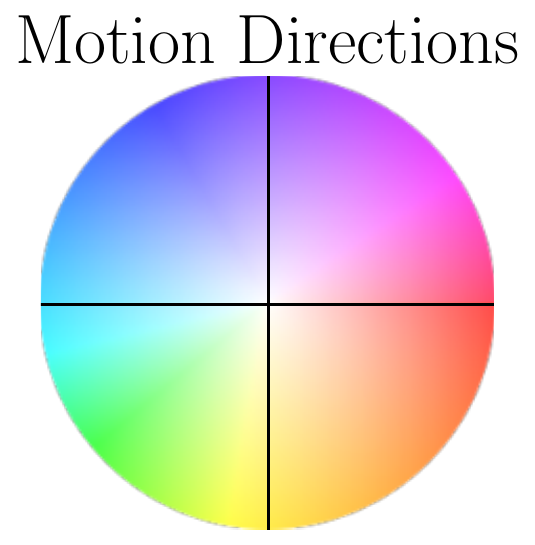}}
  \end{minipage}
\end{figure*}

\parafango{Qualitative Results} Fig.~\ref{fig:main_frame} presents a qualitative analysis of the outputs of our models on three sample frames, with similar trends observed across all video streams. 
Specifically, the second and third columns depict the extracted flows from \acronymm{}'s first and second levels, respectively. The features produced by \acronymm{} allow the classification procedure (recall that such procedure do not affect the feature learning process) to completely discriminate all the different target objects (fourth column, predictions).  While the borders of the \acronymm{}-based predictions appear slightly thicker compared to the ground-truth object borders, this is primarily due to the quality of the estimated motion field, which is not always perfect (recall that it is learned from scratch jointly with the features). In fact, contrastive learning is affected by noisy flow, which can easily result in \emph{false} positive pairs, with one pixel belonging to an object and the other just outside of it. Noticeably, the estimated optical flow (second column) is very crisp in the case of \textsc{EmptySpace} where the camera is fixed, while it captures the effects of the moving-camera in real-world streams (second and third row). Also, optical flow estimation is known to be inherently challenging in nearly texture-free surfaces. Interestingly, comparing the two flow estimators of distinct levels (second and third column), we notice how they appear to focus on different parts of the frames, with flows that are somewhat related but clearly not overlapping. This confirms that the network develop a {\it latent} higher-order motion, challenging to interpret, but well suited to the learning process.
It is worth noting that we did not use any specific tricks \cite{zhai2021optical} to improve the flow extraction, such as spatial pyramids or occlusion handling, for the sake of simplicity, consistently with \cite{marullo2022continual}.

\begin{figure}
    \centering
        \includegraphics[width=0.34\textwidth]{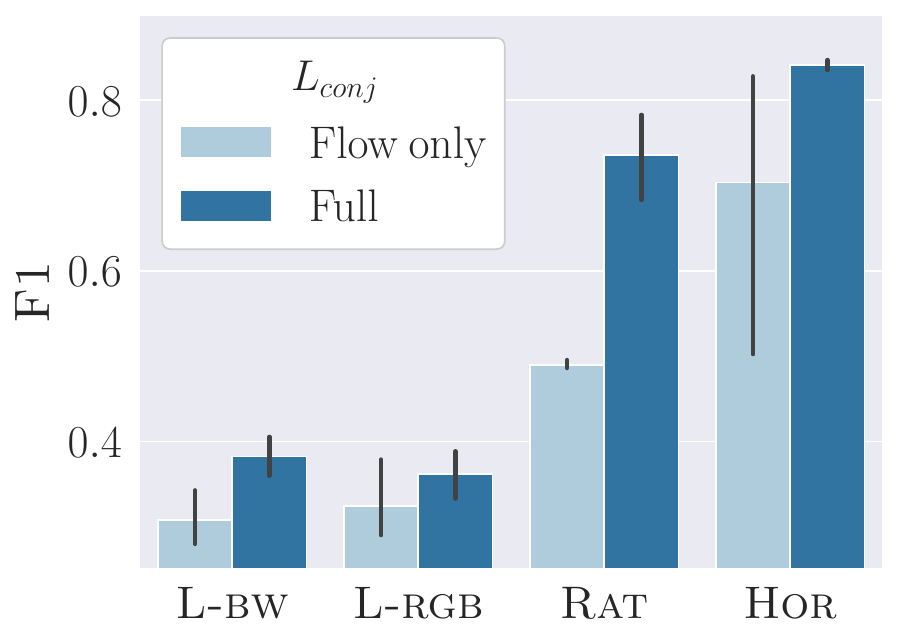}
        \includegraphics[width=0.34\textwidth]{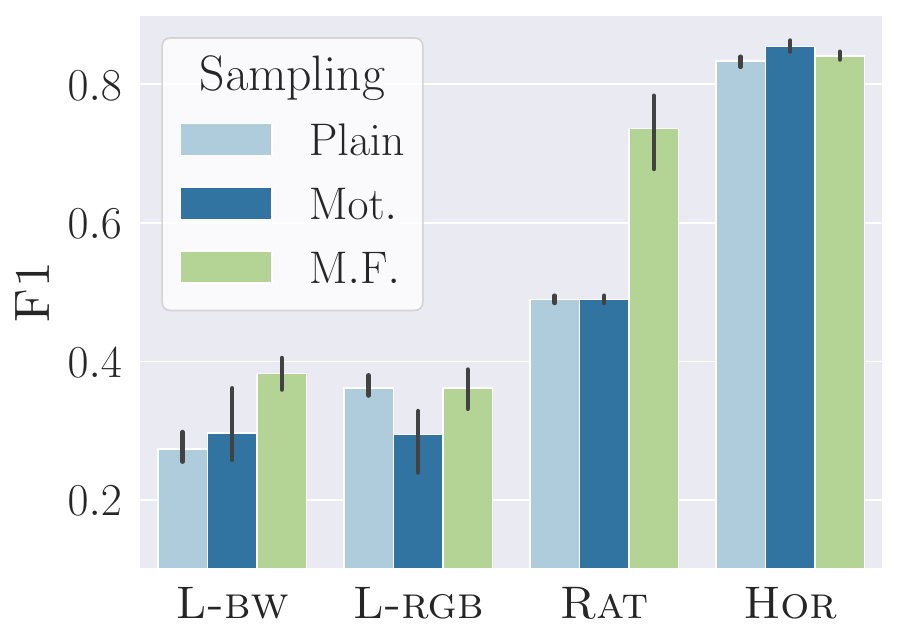}\\
        \includegraphics[width=0.34\textwidth]{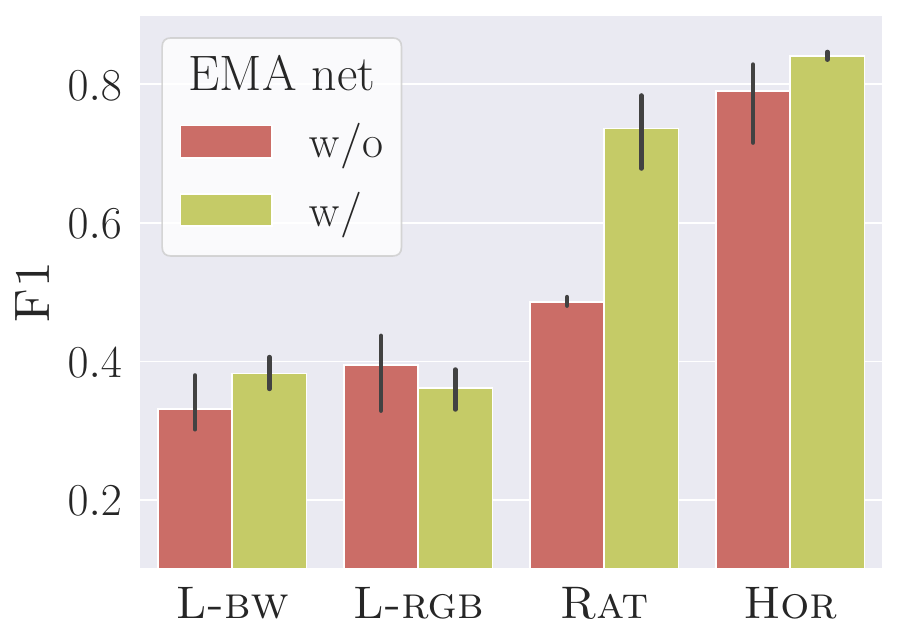}
        \includegraphics[width=0.34\textwidth]{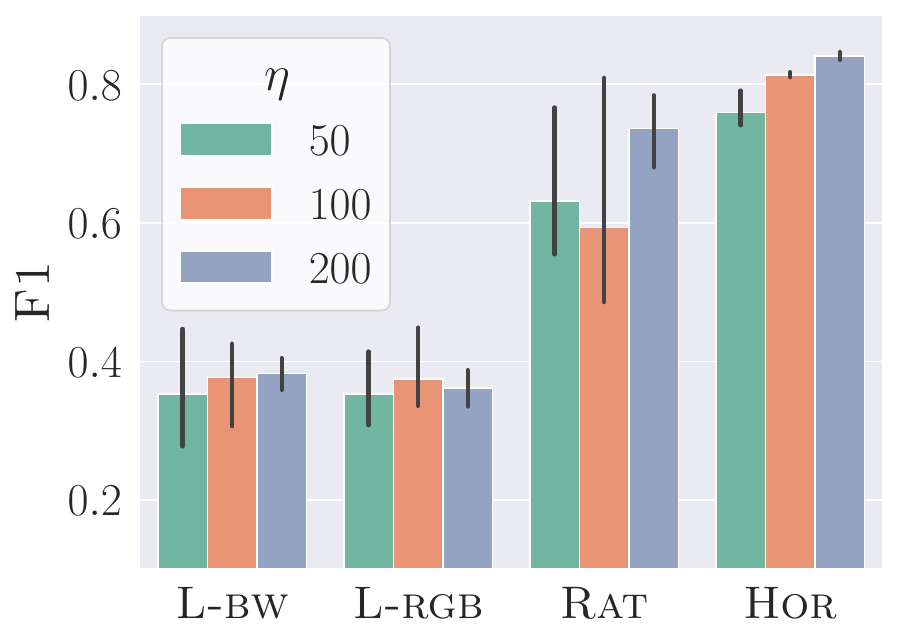}
    \caption{Ablation of the model components on four visually-rich streams (\textsc{L} stands for \textsc{LivingRoom}, \textsc{Hor} for \textsc{Horse}). From left to right, top to bottom: the impact of the features-motion conjugation term ${L}_{conj}$ as presented in Eq.~\ref{eq:megaconj} (Full) compared to the case in which features are only developed by enforcing them to be coherent with optical flow (i.e., when Eq.~\ref{eq:megaconj} is composed of $ii.$ only--Flow only); the role of different sampling strategies (Plain, Motion--Mot., Motion and Features--M.F.); EMA network (without or with); number of sampled points $\eta$.}
    \label{fig:ablations}
\end{figure}

\parafango{Ablation studies}  In order to better understand the impact of key design choices, we perform ablation experiments concerning some of the main components of \acronymm{}, reporting our findings in Fig.~\ref{fig:ablations}  that we describe from left to right, top to bottom.
The motion-feature conjugation term ${L}_{conj}$ from the loss function (Eq.~\ref{eq:megaconj}) provides improvements in all the streams ($1$st barplot), when compared to the case in which we simply enforce consistency with optical flow (i.e., Eq.~\ref{eq:megaconj} is composed of $ii.$ only), more evidently in real-world streams. It also improves stability among the different runs, resulting in reduced standard deviation---a part for \textsc{Rat}, where the lack of full conjugation leads to collapse to a very suboptimal solution with minimal variance. The adoption of a specific sampling strategy ($2$nd barplot) for the contrastive term, guided by both Motion and Features (\textrm{M.F.} in Fig.~\ref{fig:ablations}), also had a considerable influence on the results, sometimes dramatic as in the case of \textsc{Rat}. In general, we notice that different sampling strategy might be optimal in different streams, in relation with the kind of motion that is present in the respective environments. Motion-only-guided sampling (\textrm{Mot.}) can yield slight improvements w.r.t. vanilla uniform sampling strategy (\textrm{Plain}), due to the fact that it helps focusing on the rather small moving areas. 
Learning without the stabilizing effect of the EMA  network ($3$rd barplot) generally leads to lower performance, in agreement with literature in both contrastive learning and continual learning. The learning process is however still effective (a part from the challenging \textsc{Rat}), thanks to regularization effects introduced by the spatio-temporal loss terms, although less stable than when using the EMA scheme. 
 We also separately evaluated the impact of the number of sampled locations $\eta$ ($4$th barplot). It turns out that models developed in real-world streams consistently benefit from a larger number of locations ($\eta=200$). Conversely, this trend is not evident in synthetic streams, that have either large static or uniform areas, where more dense sampling does not yield more information. 



\parafango{Limitations} Recent works highlighted the inherent unstable nature of self-supervised contrastive criteria \cite{zhangdoes}.
\acronymm{} is a pixel-wise motion-based contrastive criterion, proposed in the context of a completely online learning procedure, and as such there are some stability issues, confirmed by the non-negligible standard deviation of our results over multiple different initializations (Table~\ref{tab:main_frame2}).
Another intrinsic source of instability stands on the sensibility to the effects of the first learning steps, where the flow estimators are obviously far from being able to make reasonable predictors. In fact, the proposed contrastive loss (Eq.~\ref{eq:ssl}) depends on how motion gets predicted, thus early-stage prediction might drive the feature extractor(s) toward configurations that we sometimes found to have negative effects on the following development of the model.
\acronymm{} (as well as its main competitor) is designed to learn in environments where background areas are characterized by motion patterns that are different from the ones of objects to which semantics are expected to be attached (or that are almost static). 
 Our results show that small (or relatively slow) camera motion, as the one of the selected real-world videos, can be easily handled or filtered out using a larger $\tau_{m}$ or setting it to the average length of motion vectors in the current frame, as we did in our implementation. However, dealing with strongly moving cameras might lead to development of features that encode information which is less object-specific. 

\section{Conclusions}
\label{sec:conclusions}
In this work we focused on learning to predict pixel-wise information in a fully unsupervised continual manner. We considered multi-level features and multi-level (order) motion flows as ``conjugated'' one to each other, mutually constraining their development. 
We proposed \acronymm{}, that, to our best knowledge, is the first implementation of these principles. \acronymm{} is based on neural models trained to extract features and flows from scratch, driven by multi-level spatio-temporal feature-motion constraints and on a self-supervised (contrastive) loss that is biased by the way flows are predicted. 
We performed experiments on video streams from a recent benchmark and real-world videos, showing how \acronymm{} benefits from learning multi-order flows, and how it significantly overcomes its main competitor. 
\acronymm{} also behaves similarly to larger networks that were offline-pre-trained on large collections of supervised and unsupervised images.
The novelty of the ideas in this paper are paired with limitations which we explicitly discussed, and that will be the subject of our future studies. Finally, another direction for future work is the use of \acronymm{} jointly with more advanced continual learning strategies to avoid forgetting issues in significantly longer streams or with several object categories.

%

\bibliography{biblio}{}
\bibliographystyle{plain}

\begin{appendices}
\section*{Appendix}

This appendix provides further details about the proposed model. This includes in-depth descriptions of the neural architectures and the selected competitors, the grids of values of the hyper-parameters that we validated in our experimental activity, together with the best selected values. We also provide additional results and qualitative investigations, as well as further information on how to reproduce the results of this paper (including implementation-oriented descriptions). 
Notice that the bibliographic references contained in this appendix are about the bibliography of the main paper.

\section{Neural Architectures}
\label{sec:models}
We provide additional details on the selected neural architectures and on the competitors.

\parafango{Feature Extractors}  The neural branch of \acronym{} yielding features, i.e., function $f$, is a ResUnet architecture, inspired to the one presented in \cite{ijcaistocazzic} but having a smaller amount of parameters. It is a UNet-like \cite{ronneberger2015u} architecture\footnote{Inspired to \url{https://github.com/usuyama/pytorch-unet} (MIT License), but with one fourth of filters in each layer. Also, the last layer has an output dimensionality of 32 features}, based on a ResNet18 backbone.  
Among the pretrained competitors, DPT-Hybrid \cite{ranftl2021vision} is a Transformer-based ViT model, trained on the ADE20K dataset \cite{zhou2019semantic}. All the other considered models adopt ResNet backbones, in particular the ResNet-50 architecture, except for DeepLab v3 (ResNet-101). DeepLab models apply Atrous Spatial Pyramid Pooling (ASPP) on top of the backbone features and they are trained on COCO dataset \cite{lin2014microsoft}\footnote{Available at \url{https://pytorch.org/vision/stable/models.html}}.
\textsc{MoCo} models\footnote{Available at \url{https://github.com/facebookresearch/moco}} are trained with a contrastive unsupervised objective on ImageNet-1M. \textsc{MoCo v2}  differs from the original version because of an improved training procedure (better augmentation, cosine learning rate schedule, extra MLP in training phase). 
\textsc{MoCo v3}\footnote{Available at \url{https://github.com/facebookresearch/moco-v3}} introduces some tricks to improve the learning stability both for ViT and Resnets backbones. We focus our analysis on the latter backbone. 
In the case of PixPro\footnote{Available at \url{https://github.com/zdaxie/PixPro}} models, we use backbone weights obtained from pretraining on ImageNet-1M with their pixel-level contrastive task.

\parafango{Motion Estimation} \acronymm{} neural branch for motion estimation, i.e., function $\delta$, inherits its structure and composition from the feature extraction one (ResUnet), except that we cut the skip connections connecting the input to the output layer, since we empyrically noticed in preliminary experiments that we could obtain a more sensible estimation. Such a network outputs the vertical and horizontal displacement maps for all the frame pixels. 

\section{Additional implementation details and hyper-parameters}
\label{sec:params}
In order to complement the already described experimental setup of Section~\ref{sec:exp}, here we introduce some further details to make our experience reproducible.
For completeness we report that, in our implementation, the maximum distance between pairs of pixels in Eq.~\ref{eq:posnegpn} (i.e., $\sqrt{H^2+W^2}$ in the normalization factors) is replaced with the maximum distance between pairs of sampled coordinates. 

\parafango{Data augmentation} For each processed frame pair, we create a mini-batch composed of the two original frames $({I}_{t-1}, {I}_{t})$, as well as multiple augmented views/transformations, in order to improve the feature robustness, as common in deep net training procedures.
Augmentations are of 3 different types: (A) large random crops (with crop ratio greater than 0.9) followed by upscaling to $W \times H$; (B) random horizontal/vertical flips; (C) random color distortion and Gaussian blur. 
For the augmentations of type (A) and (C), only one of the frames composing the pair $({I}_{t-1}, {I}_{t})$ is transformed, randomly selected at each $t$, whilst the other is kept as it is. 

\parafango{Parameters and validation} In order to make the comparisons fair, we followed the exact same hyper-parameter validation procedure of \cite{ijcaistocazzic}. In particular, for each video 
stream we selected the values of the hyper-parameters which maximize the F1 score measured during the 30-th lap, only considering one pixel per frame, along the trajectory of a human-like focus of attention mechanism attending the scene. We sampled the best hyper-parameters from the following grids: $\lambda_m \in \{1\}$, $\beta_m \in \{10^{-4}, 10^{-3}, 10^{-2},\}$, $\beta_f \in \{1\}$,  $\tau_p \in \{0.7, 0.8, 0.9\}$,  $\tau_n \in \{-0.5, -0.3, 0, 0.3, 0.5  \}$, $\ell\in \{50, 100, 200\}$,  $\tau \in \{0.1, 0.5\}$,  $\tau_m \in \{0.5, 1.0, 1.5, 2.0\}$, $\alpha_m \in  \{10^{-4}, 5 \cdot 10^{-4}, 10^{-3}, 10^{-2}\}$, $\alpha_f \in  \{10^{-4}, 5 \cdot 10^{-4}, 10^{-3}, 10^{-2}\}$, $\xi \in  \{0.5, 0.9, 0.99\}$.  

An additional speed-up of the training process can be achieved by further limiting the summations of Eq.~\ref{eq:fava} (both of them) by keeping only a fraction $\aleph \in (0,1]$ of the positive and negative pairs (balanced). In particular, we experimented the case in which we kept those positive (resp. negative) pairs for which representations are the least similar (resp. most similar), i.e., focusing on the ones that mostly contribute to larger values of $L_{self}$. In the hyperparameter search, we consider $\aleph \in \{0.5, 0.7, 1.0\}$.
The motion estimator $\delta$ is optimized using Adam \cite{kingma2014adam}, while we empirically found better results by optimizing the features extractor $f$ with SGD (with the exception of the \textsc{EmptySpace} stream, where Adam usually yields better performances). In addition to the typical smoothness regularizer introduced in Eq.~\ref{eq:megaconj}, we also exploited a further regularization modulated by a customizable parameter $\lambda_{r}$, in order to keep the estimated motion magnitude small in the case of uniform backgrounds,
\begin{equation}
    \mathcal{R}(\delta^\ell_t) = \lambda_s \cdot ({wh})^{-1} \sum_{x \in \mathcal{X}}\| \nabla\delta_t^{\ell}(x) \|^2 + \lambda_r \cdot ({wh})^{-1} \sum_{x \in \mathcal{X}}\| \delta_t^{\ell}(x) \|^2 
\end{equation}
with $\lambda_r, \lambda_{s} \in \{10^{-4}, 5\times 10^{-4}, 10^{-3}, 5\times10^{-3}, 10^{-2}\}$. 
With reference to Eq.~\ref{eq:megaconj}, in the implementation we have appropriate coefficients for the three terms:

\begin{equation}
\begin{aligned}
{L}_{conj}^{\ell} = &
\lambda_{cur}^{\ell} \mathcal{L}_{c}({\delta_{t}^{\ell}}, f_{t-1}^{\ell}, f_{t}^{\ell}) \\ + & \lambda_{skip}^{\ell}\mathcal{L}_{c}({\delta_{t}^{1}}, f_{t-1}^{\ell}, f_{t}^{\ell}) \\ +& \lambda_{low}^{\ell}\mathcal{L}_{c}(\delta_{t}^{\ell}, f_{t-1}^{\ell-1}, f_{t}^{\ell-1}) \\ +& \mathcal{R}(\delta^\ell_t)
\end{aligned}
\end{equation}
with the abovementioned coefficients in the range  $\{1, 2\} \times \{10^{-5}, 10^{-4}, 10^{-3}, \\ 10^{-2}, 10^{-1}\}$. Concerning the learning schedule $T_{sched}$ (motion estimator warmup), we tried the following values $\{0, 2, 10\}$ laps, where $0$ means no scheduling (the entire model is updated starting on the very first frame). 
In all our experiments, we reported the average results obtained over 10 runs, initializing random generators with seeds (\texttt{torch.manual\_seed} in PyTorch) from the following list: $\{$1234, 123456,
       12345678,
        1234567890,
       9876,
       98765,
       987654,
        9876543,
       98765432,
        987654321$\}$.

We report in Tab.~\ref{tab:best_params} the best-found values for the hyperparameters. 

\setlength{\tabcolsep}{2pt}
\begin{table}[t]
\centering
\footnotesize
\caption{Optimal parameters. The best selected hyperparameters drawn from the grids described in the text, for all the datasets. See the code for further details.
 }\label{tab:best_params}
\begin{tabular}{lc|c@{\hspace{1mm}}c@{\hspace{1mm}}c@{\hspace{1mm}}c@{\hspace{1mm}}c@{\hspace{1mm}}c@{\hspace{1mm}}c}
\toprule
&  & \multicolumn{2}{c}{\textsc{EmptySpace}} &  \textsc{Solid}  & \multicolumn{2}{c}{\textsc{LivingRoom}}&  \textsc{Rat} &  \textsc{Horse} \\
& Parameters & {\tiny BW} & {\tiny RGB} & {\tiny BW} & {\tiny BW} & {\tiny RGB} & {\tiny RGB}& {\tiny RGB} \\
\toprule
& $\aleph$ & 0.7 & 1.0 & 1.0 & 1.0 & 1.0 & 0.7 & 1.0\\
& $\alpha_f$ & $10^{-3}$ & $10^{-3}$ & $10^{-2}$ & $10^{-3}$ & $10^{-3}$ & $10^{-3}$ & $10^{-4}$\\
& $\alpha_m$ & $10^{-4}$ & $10^{-4}$ & $10^{-4}$ & $10^{-4}$ & $10^{-4}$ & 5$\times10^{-4}$ & $10^{-4}$ \\
& $\eta$ & 100 & 100 & 100 & 200 & 200 & 200 & 200\\
& $\lambda_{low}^0$ & 1 & 1 & 1 & 1 & 1 & 1 & 1\\
& $\lambda_{low}^1$ & 0.1 & 1 & 1 & 1 & 1 & 0.01 & 0.01 \\
& $\lambda_{cur}^0$ & $10^{-4}$ & $10^{-4}$ & $10^{-2}$ & $10^{-4}$ & $10^{-4}$ & $10^{-4}$ & $10^{-2}$ \\
& $\lambda_{cur}^1$ & $10^{-2}$ & $10^{-4}$ & $10^{-3}$ & $10^{-2}$ & $10^{-2}$ & $10^{-4}$ & $10^{-2}$\\
& $\lambda_{skip}^0$ & 0 & 0 & 0 & 0.2 & 0.2 & 2$\times10^{-5}$ & 2$\times10^{-3}$ \\
& $\lambda_{skip}^1$ & 0 & 0 & 0 & 0.2 & 0.2 & 2$\times10^{-5}$ & 2$\times10^{-3}$\\
& $\tau$ & 0.5 & 0.1 & 0.5 & 0.1 & 0.1 & 0.5 & 0.5\\
& $\tau_m$ & 1.5 & 1.5 & 2 & 0.7 & 0.7 & 1.3 & 1.3\\
& $\tau_n$ & 0 & 0 & -0.3 & -0.5 & -0.5 & 0.5 & 0.5\\
& $\tau_p$ & 0.9 & 0.7 & 0.7 & 0.9 & 0.9 & 0.8 & 0.8\\
& $\xi$ & 0.99 & 0.5 & 0.99 & 0.99 & 0.99 &0.99 & 0.99\\
& $\lambda_{r}$ & $10^{-3}$ & $5\times10^{-3}$ & $10^{-3}$ & $10^{-2}$ & $10^{-2}$ & $10^{-2}$ & $10^{-2}$\\
& $\lambda_{s}$ & $10^{-4}$ & $10^{-4}$ & $10^{-4}$ & $10^{-4}$ & $10^{-4}$ & $10^{-3}$ & $10^{-4}$\\
& $T_{sched}$ & $10$ & $0$ & $0$ & $0$ & $0$ & $10$ & $2$\\

\bottomrule 
\end{tabular}
\end{table}

\section{Additional Qualitative Analysis}
\label{addqual}

\begin{figure*}[!ht]
    \scriptsize
    \centering
     \hspace{-.121\textwidth} \hskip 0.5mm \rotatebox{90}{\hskip 7mm \textsc{Frame}} 
     \fbox{\includegraphics[width=.13\textwidth]{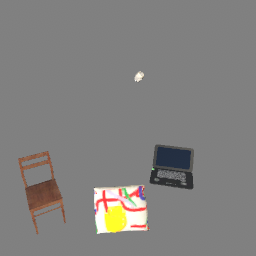}}
     \fbox{\includegraphics[width=.13\textwidth]{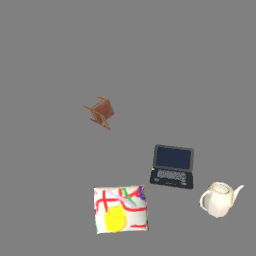}}
     \fbox{\includegraphics[width=.13\textwidth]{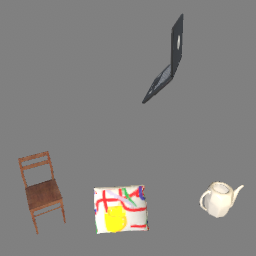}}
     \fbox{\includegraphics[width=.13\textwidth]{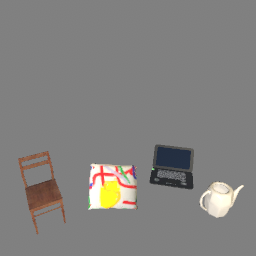}}
     \fbox{\includegraphics[width=.13\textwidth]{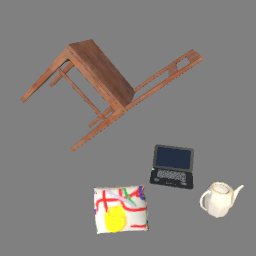}}
     \fbox{\includegraphics[width=.13\textwidth]{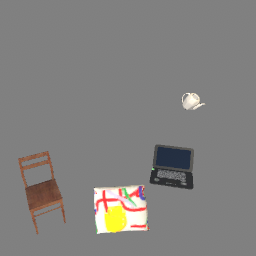}}\\
      \hskip 0.8mm
     \rotatebox{90}{\hskip 3mm Tiezzi et al. \cite{ijcaistocazzic}}
     \fbox{\includegraphics[width=.13\textwidth]{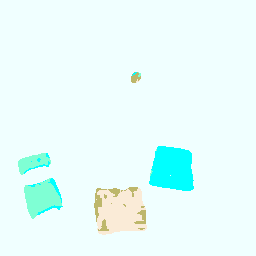}}
     \fbox{\includegraphics[width=.13\textwidth]{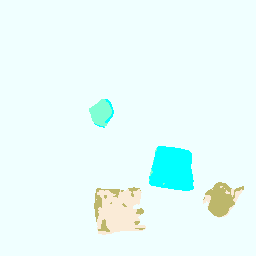}}
     \fbox{\includegraphics[width=.13\textwidth]{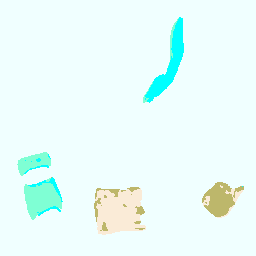}}
     \fbox{\includegraphics[width=.13\textwidth]{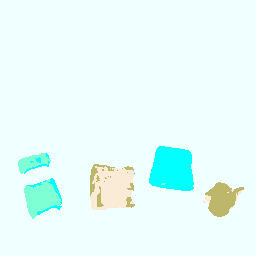}}
     \fbox{\includegraphics[width=.13\textwidth]{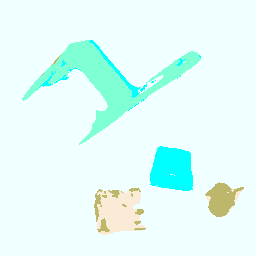}}
     \fbox{\includegraphics[width=.13\textwidth]{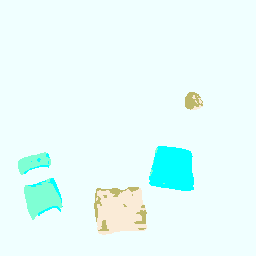}}
     \raisebox{.05\height}{\includegraphics[width=.115\textwidth]{fig/seg_legend.pdf}}\\
     \hspace{-.121\textwidth} \hskip 0.5mm \rotatebox{90}{\hskip 7mm \acronymm}
     \fbox{\includegraphics[width=.13\textwidth]{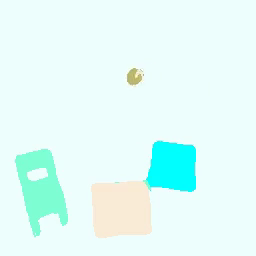}}
     \fbox{\includegraphics[width=.13\textwidth]{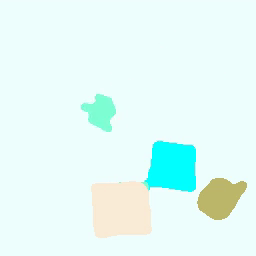}}
     \fbox{\includegraphics[width=.13\textwidth]{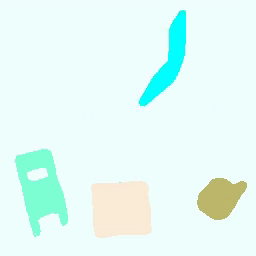}}
     \fbox{\includegraphics[width=.13\textwidth]{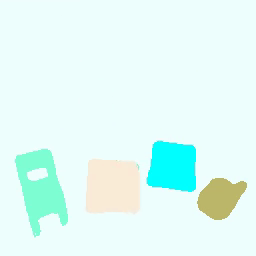}}
     \fbox{\includegraphics[width=.13\textwidth]{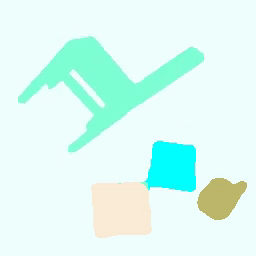}}
     \fbox{\includegraphics[width=.13\textwidth]{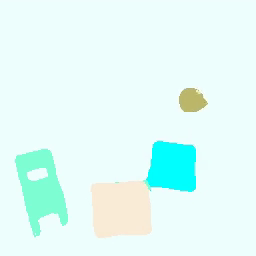}}\\
     
    \caption{Six frames sampled from the \textsc{EmptySpace-rgb} stream (first row). We compare the predictions yielded by the main competitor (second row) and the outputs by \acronymm{} (third row). Different colors are used to denote different predictions.}
    \label{fig:predictions2}
\end{figure*}
We report in Fig.~\ref{fig:predictions2} some additional qualitative results of the predictions obtained leveraging the features by \acronymm{}, on six frames from the \textsc{EmptySpace-rgb} stream (first row). 
The model by \cite{ijcaistocazzic} yields features which are not capable to correctly discriminate pixels from the chair legs/thinner segments, as long as some textures on the pillow: see for instance the borders of the pillow which are misled with the ewer (given the similar brightness/RGB values). Similarly, some parts of the ewer (the spout and the handle) are mistaken for pixels belonging to the pillow. Conversely,  \acronymm{} allows the classification procedure (which is not-involved in the feature learning process) to almost completely disentangle all the different objects and their parts, even when they appear in peculiar poses and orientations (see the chair legs in the second and fifth image). As a general consideration, CMOSS yields features (and thus predictions) which tend to slightly overflow with respect to the ground-truth object borders. 

\section{Code}
\label{code}
We provide a Python implementation of \acronymm{}, as well as instructions to run experiments \footnote{ \url{https://github.com/sailab-code/unsupervised-learning-feature-flow}}. 
The \textsc{README} file contained in the archive describes how to execute the code, and it gives further details on the mapping between the notation used in the main paper for naming hyperparameters and the one used in the code (see ``RUNNING EXPERIMENTS - Argument description/mapping with respect to the paper notation'' therein).

\end{appendices}
\end{document}